\documentclass{article}

\usepackage{arxiv}

\usepackage[utf8]{inputenc}
\usepackage{amsmath}
\usepackage{mathrsfs}
\usepackage{amsmath,bm}
\usepackage{amssymb}
\usepackage{bbm}
\usepackage{graphicx}
\usepackage{esint}
\usepackage{enumitem}
\usepackage{amsthm}
\usepackage{amsfonts}
\usepackage{booktabs}
\usepackage{caption,subcaption}
\usepackage{epstopdf}
\usepackage[svgnames]{xcolor}
\usepackage{listings}
\usepackage{mathtools}
\usepackage{authblk}
\usepackage[ruled,vlined]{algorithm2e}
\usepackage[colorlinks=true,linkcolor=blue,citecolor=blue]{hyperref}
\usepackage[sort, numbers]{natbib}
\usepackage{tikz}
\usetikzlibrary{shapes,arrows,positioning,fit,calc}
\tikzset{block/.style={draw, thick, text width=3cm, minimum height=1.5cm, align=center},   
line/.style={-latex}   
}  
\tikzstyle{block} = [rectangle, draw, 
    text width=5em, text centered, rounded corners, minimum height=4em]
\tikzstyle{line} = [draw, -latex']

\newcommand{\E}{\mathbb{E}}
\newcommand{\R}{\mathbb{R}}
\newcommand{\N}{\mathcal{N}}

\newcommand{\ds}{\displaystyle}

\newcommand{\df}{\text{d}}

\SetKwInput{kwInit}{initilize}
\title{Variational Inference  with NoFAS:\\ Normalizing Flow with  Adaptive Surrogate for Computationally Expensive Models}
\author[]{Yu Wang}
\author[]{Fang Liu}
\author[]{Daniele E. Schiavazzi}
\affil[]{Department of Applied and Computational Mathematics and Statistics, University of Notre Dame, Notre Dame, IN, USA}
\date{ }

\begin{document}
\maketitle


\begin{abstract}
\noindent Fast inference of numerical model parameters from data is an important prerequisite to generate predictive models for a wide range of applications.
Use of sampling-based approaches such as Markov chain Monte Carlo may become intractable when each likelihood evaluation is computationally expensive.
New approaches combining variational inference with normalizing flow are characterized by a computational cost that grows only linearly with the dimensionality of the latent variable space, and rely on gradient-based optimization instead of sampling, providing a more efficient approach for Bayesian inference about the model parameters.
Moreover, the cost of frequently evaluating an expensive likelihood can be mitigated by replacing the true model with an offline trained surrogate model, such as neural networks. However, this approach might generate significant bias when the surrogate is insufficiently accurate around the posterior modes.
To reduce the computational cost without sacrificing inferential accuracy, we propose Normalizing Flow with Adaptive Surrogate (NoFAS), an optimization strategy that alternatively updates the normalizing flow parameters and surrogate model parameters. We also propose an efficient sample weighting scheme for surrogate model training that preserves global accuracy while effectively capturing high posterior density regions.
We demonstrate the inferential and computational superiority of NoFAS against various benchmarks, including cases where the underlying model lacks identifiability. The source code and numerical experiments used for this study are available at \url{https://github.com/cedricwangyu/NoFAS}.
\end{abstract}

\section{Introduction}
%
\noindent Numerical models are increasingly used to improve our understanding of physical processes in engineering and science.
They can achieve a remarkable realism, even in the simulation of complex multi-physics phenomena, but their computational cost typically increases with their complexity, and with the level of detail they are designed to provide. These models might, in addition, contain a large number of parameters that need to be carefully tuned to reproduce observed data, before \emph{predicting} the response of the system of interest for unobserved conditions.

This two-step process of first inferring the model parameters from data and then using an optimally trained model to make predictions is essential to construct \emph{digital twins}, i.e., predictive digital replicas of physical systems. 
Recent applications combining digital twins with Bayesian network-based probabilistic reasoning are discussed, for example, in~\cite{kapteyn2021probabilistic} for damage assessment in unmanned aerial vehicles, and~\cite{schiavazzi2020bayesian} for multi-physics simulations for hypersonic systems. A recent digital twin for computational physiology is also discussed in~\cite{harrod2021predictive} for predicting group II pulmonary hypertension in adults.

The most expensive computational task in the development of predictive models is the solution of an inverse problem. Since the exact posterior distribution of the model parameters given a set of observations is not available in closed form, posterior sampling is often employed to approximate the posterior distribution numerically, using approaches such as Approximate Bayesian Computation (ABC)~\cite{beaumont2019approximate} and, in the presence of a tractable likelihood, Markov chain Monte Carlo sampling (MCMC)~\cite{gilks1995markov}. 
For complicated posteriors with multiple modes or ridges, advanced MCMC with adaptive proposal distribution were proposed in~\cite{haario2001adaptive,haario2006dram}, with more recent approaches being the Hamiltonian Monte Carlo~\cite{neal2011mcmc}, the no U-turn sampler~\cite{hoffman2014no} and the differential evolution adaptive Metropolis (DREAM) algorithm~\cite{vrugt2009accelerating,vrugt2016markov} to cite a few.

As an alternative to sampling-based approaches, variational inference (VI)~\cite{jordan1999introduction,wainwright2008graphical,blei2017variational} leverages numerical optimization to determine the member of a parametric family of distributions that is the closest to a desired posterior in some sense (e.g., the Kullback-Leibler divergence). 
VI can be combined with stochastic optimization to improve its efficiency for large datasets~\cite{hoffman2013stochastic}, but still required significant model-specific analysis. To overcome this problem, Black-Box VI (BBVI~\cite{ranganath2014black}) was developed to support a larger class of models, leveraging Monte Carlo estimates of the evidence lower bound (ELBO) and reparameterized gradients with lower variance~\cite{salimans2013fixed,kingma2013auto,rezende2014stochastic,ruiz2016generalized}.
However, a common approximation made by BBVI is the mean field assumption, which enforces an a-priori independence among groups of parameters, and may introduce significant bias in the distributional approximation, when the independence assumption does not hold. To go beyond the mean field assumption, many approaches have been proposed using, e.g., Gaussian mixture models~\cite{blei2017variational}, hierarchical variational models~\cite{ranganath2016hierarchical}, CRVI~\cite{fazelnia2018crvi}, copula VI~\cite{tran2015copula} and others (see, e.g.,~\cite{zhang2018advances}), but they often rely on strong parametric assumption when introducing a dependency structure. 

Normalizing flows was proposed in~\cite{rezende2015variational}. It can be used for VI via a composition of invertible transformations with easily computable Jacobian determinants. Simple flows were initially proposed such as planar and radial flows~\cite{rezende2015variational}, followed by Inverse Autoregressive Flow (IAF~\cite{kingma2016improved}), real-valued non volume preserving transformations (RealNVP~\cite{dinh2016density}), masked autoregressive flow (MAF~\cite{papamakarios2017masked}), generative flow (GLOW~\cite{kingma2018glow}), among others. Readers may refer to~\cite{kobyzev2020normalizing} for an extensive overview on NF.

For both sampling-based posterior inference or approximate inference via VI through NF, the likelihood function given the assumed model and data need to be repeatedly evaluated for new posterior samples, which becomes quickly infeasible, in practice, for computationally expensive models. 
One of the solutions is to approximate the models using a surrogate $\hat{f}$ which is generally obtained through a three-phase process, i.e., input dimensionality reduction, design of experiments, and formulation of a surrogate from an appropriate family (see, e.g.,~\cite{alizadeh2020managing}).
Reduction of the inputs dimensionality can be achieved through, e.g., principal component analysis~\cite{gao2018categorical}, variable screening~\cite{cho2014efficient, koch1999statistical, bettonvil1997searching}, sensitivity analysis~\cite{helton1999uncertainty, sobol1993sensitivity, rabitz1989systems}, and Bayesian updating~\cite{beck2002bayesian, kennedy2001bayesian}. 
Sampling locations can be selected through factorial design~\cite{gunst2009fractional}, central composite design~\cite{montgomery2017design} and orthogonal arrays~\cite{hedayat2012orthogonal}, among others.
Finally, possible surrogate formulations include response surface analysis~\cite{khuri2010response}, Kriging~\cite{stein2012interpolation}, radial basis functions~\cite{shan2010survey}, boosting trees and random forests~\cite{friedman2001greedy}, adaptive and active learning \cite{carbonell1970ai, gorissen2010surrogate}, among others.
Popular non-intrusive approaches used in computational mechanics include the generalized polynomial chaos expansion (gPC)~\cite{xiu2002wiener, ernst2012convergence}, stochastic collocation on tensor isotropic and anisotropic quadrature grids~\cite{babuvska2007stochastic, nobile2008anisotropic}, multi-element adaptive gPC~\cite{wan2005adaptive}, hierarchical sparse grids~\cite{ma2009adaptive}, simplex stochastic collocation~\cite{witteveen2012simplex}, sparsity-promoting gPC~\cite{doostan2011non}, generalized multi-resolution expansion~\cite{schiavazzi2014sparse, schiavazzi2017generalized} and, more recently, deep neural networks~\cite{tripathy2018deep}.
Finally, other approaches in the literature have proposed the combination of an inference task and an adaptive surrogate model. For example, a local approximant is combined with MCMC in~\cite{conrad2016accelerating, conrad2018parallel, davis2020rate} and a Gaussian process surrogate coupled with Hamiltonian Monte Carlo sampling is presented in~\cite{paun2021mcmcgp} with applications to one-dimensional hemodynamics.

In this work, we combine NF and surrogate modelling and propose a new method -- Normalizing Flow with Adaptive Surrogate (NoFAS) -- for variational inference with computationally expensive models. NoFAS is a general approach for Bayesian inference and uncertainty quantification, designed to sample from complex or high-dimensional posterior distributions with significantly reduced computational cost. 
This is an alternated optimization algorithm, where a surrogate is adaptively refined using posterior samples which are, in turn, computed by optimizing the NF parameters with respect to a loss function that depends on the surrogate.
Any expressive NFs can be used, and we find the autoregressive NFs to be effective in approximating even complicated posterior distributions. 
Our contributions are listed below.
\vspace{-5pt}
\begin{itemize}\itemsep -3pt
\item We combine variational inference, normalizing flow and surrogate modelling in a data efficient and computationally affordable framework to obtain inference on true model parameters.
\item We propose an efficient sample weighting scheme for the loss function that remembers samples in the pre-grid (providing global surrogate accuracy) as well as more recent samples acquired during the NF iterations. Older samples acquired during the early stages of VI are instead progressively forgotten. 
\item We demonstrate the application of NoFAS in multiple experiments with different types of models, and showcase its advantages over the fixed-surrogate model approach, MCMC and BBVI with the mean-field assumption, for indentifiable and non-identifiable models.
\end{itemize}

The paper is organized as follows. In Section~\ref{sec:method}, we review normalizing flow, surrogate modeling and present our NoFAS approach. In Section~\ref{sec:exp}, we apply NoFAS for the solution of inverse problems in four numerical experiments, and compare its performance in posterior inference to an approach where the surrogate is kept fixed, MH, and BBVI. We provide some discussions and final remarks in Section~\ref{sec:disucssion}.
\begin{table}[!ht]
\centering
\caption{List of mathematical symbols.}\label{tab:notation}
\resizebox{\textwidth}{!}{
\begin{tabular}{l l | l l}
\toprule
Symbol  & Description & Symbol  & Description \\
\midrule
$b$ & Batch size for NF. & $M$ & Total size of the adaptive\\
&&&  samples stored.\\
$b_1,\dots,b_{L+1}$ & Bias terms in MADE. & $n$ & Total number of observations.\\
$\beta_0$ & Pre-grid weight factor. & $p$ & Target density function.\\
$\beta_1$ & Memory decay factor. & $\pi(\cdot)$ & Prior distribution.\\
&  samples in loss.\\
$c$ & Calibration frequency. & $q_k(\bm z_k)$ & Density function of $\bm z_k$.\\
$d$ & Latent space dimensionality. & $\varphi_F, \varphi_S$ & Optimizers for NF and\\
&&&  surrogate update.\\
$D(\cdot \| \cdot)$ & KL-Divergence. & $\sigma_j,\,j=1,\cdots,m$ & Prescribed standard deviations\\ 
& & &  in log-likelihood.\\
$f_{\mu_i}, f_{\alpha_i}$ & MADE networks in MAF. & $\sigma$ & Soft-max activation.\\
$f$ & True model. & $S_0$ & Pre-grid size.\\
$\hat f$ & Surrogate model. & $T_F$ & Total number of NF iterations.\\
$F_k, F$ & NF bijections and their & $T_S$ & Total number of surrogate update \\
&composition. && iterations.\\
$h_{1},\dots,h_{L+1}$ & Activation functions in MAF. & $V, M^V, W, M^W$ & MAF: Mask matrices used in MAF.\\
$G$ & Queue for sample storage  & $\bm x, X$ & Observations and and their space.\\
& for calibration.\\
$k, K$ & Current and total number  & $\bm z, Z$ & Latent variables and their space.\\
& of NF layers. & $\bm y$ & Variables in hidden MADE layer.\\
$\ell(\bm z ; f, \bm x)$ & Likelihood function. & $Z_G$ & Adaptively selected\\
 &&& training set for $\hat f$\\
$\lambda, \lambda_k, \Lambda, \Lambda_k$ & NF parameters and their space. & $Z_P$ & Pre-grid.\\
$L_{j}(\cdot)$ & Loss function. & $\{z_K^{(i)}\}_{i=1}^b$ & Batch samples.\\
$m^{i}(k)$ & functions returning an integer & $\bm z_k$ & Output from the $k$-th NF layer.\\
 &  from $1$ to $d$.\\
$m$ & Output space dimensionality, & $\omega^*$ & Optimal surrogate model parameters.\\
& $\bm x\in\mathbb{R}^{m}$.  &  & \\
\bottomrule
\end{tabular}}
\end{table}

\section{Method}\label{sec:method}

\noindent In this section, we first review the framework of NF for VI, focusing on MAF, and RealNVP in Section~\ref{sec:nf}. We then discuss a surrogate model formulation for computationally expensive models in Section~\ref{sec:surrogate}. 
We introduce NoFAS in Section~\ref{sec:ourAlgorithm}, along with its algorithm. 
The notation introduced in this section is summarized in Table~\ref{tab:notation}.

\subsection{Autoregressive Normalizing Flow}\label{sec:nf}

\noindent NF uses the map $F: \R^d \times \Lambda\to \R^d$ with parameters $\lambda \in \Lambda$ to transform realizations from an easy-to-sample distribution, such as $\bm z_0 \sim \N(\bm{0}, I_d)$, to samples from a desired \emph{target} distribution.

Specifically, $F$ is obtained as a composition of $K$ bijections $F_k: \R^d \times \Lambda_k\to \R^d$, each parameterized by $\lambda_k \in \Lambda_k$: $F(\bm z_0; \lambda) = \left[F_K( \ \cdot \ ; \lambda_K)\circ F_{K-1}( \ \cdot \ ; \lambda_{K-1})\circ \cdots \circ F_1( \ \cdot  \ ;\lambda_1)\right](\bm z_0)$, where $\bm z_k = F_k(\bm z_{k-1}; \lambda_k)$ for $k=1,\ldots,K$.
Since $F_k( \ \cdot \ ;\lambda_k)$ is a bijection from $\bm z_{k-1}$ to $\bm z_{k}$, $q_k(\bm z_k)$, the distribution of $\bm z_k$, can be obtained by the change of variable
\begin{equation}\label{equ:changeOfVar}
q_k(\bm z_k) = q_{k-1}(\bm z_{k-1})\left|\det \frac{\partial F_k^{-1}}{\partial \bm z_{k-1}}\right| = q_{k-1}(\bm z_{k-1})\left|\det \frac{\partial F_k}{\partial \bm z_{k-1}}\right|^{-1}.
\end{equation}
Taking the logarithm and summing over $k$, Eqn.~\eqref{equ:changeOfVar} becomes
\begin{equation}
\log q_K(\bm z_K) = \log q_0(\bm z_0) - \sum_{k=1}^K \log \left|\det \frac{\partial F_k}{\partial \bm z_{k-1}}\right|.
\end{equation}
The goal is to determine an \emph{optimal} set of parameters $\lambda^{*} = (\lambda^{*}_1,\lambda^{*}_2, \cdots, \lambda^{*}_K) \in \Lambda_1 \times \Lambda_2\times \cdots \times \Lambda_K = \Lambda$ so the density $q_{K}$ can approximate a target density $p$. A commonly used objective (loss) function to achieve this goal is the \textit{flow-based free energy bound}, expressed as
\begin{equation}\label{equ:klDiv}
\begin{matrix*}[l]
\mathcal{F}(\bm x) & \ds = \E_{q_K(\bm z)}\left[\log q_K(\bm z) - \log p(\bm x, \bm z)\right] = \E_{q_0(\bm z_0)}\left[\log q_K(\bm z_K) - \log p(\bm x, \bm z_K)\right]\\
& \ds = \E_{q_0(\bm z_0)}[\log q_0(\bm z_0)] - \E_{q_0(\bm z_0)}[\log p(\bm x, \bm z_K)] - \E_{q_0(\bm z_0)}\left[\sum_{k=1}^K \log \left|\det \frac{\partial F_k}{\partial \bm z_{k-1}}\right|\right].
\end{matrix*}
\end{equation}
The expectations in Eqn.~\eqref{equ:klDiv} are approximated by their Monte-Carlo (MC) estimates using samples $\bm z_0$ from the basic distribution $q_0$.
However, the computation of the Jacobian determinants in Eqn.~\eqref{equ:klDiv} may be computationally intensive, especially when using a large number of bijections. 
To efficiently compute the determinants, the coupling layer-based NF with block-triangular Jacobian matrices (RealNVP~\cite{dinh2016density} and GLOW~\cite{kingma2018glow}), and autoregressive transformation-based NF with lower-triangular Jacobian matrices (MAF~\cite{papamakarios2017masked} and IAF~\cite{kingma2016improved}) have been proposed. In this study, we focus on autoregressive transformations in NF.

According to the chain rule, the joint distribution $p(\bm z)$ can be written as $p(\bm z) =$ $p_1(z_1)$ $\prod_{i=2}^d p_i(z_i|z_{1},\dots,z_{i-1})$. If the components of $\bm z$ are not independent, the autoregressive flow can be applied to capture their dependency.
For example, MAF~\cite{papamakarios2017masked} uses $p(z_i|z_{1},\dots,z_{i-1}) = \phi((z_i - \mu_i) / e^{\alpha_i})$, where $\phi$ is the density function of the standard normal distribution, $\mu_i = f_{\mu_i}(z_{1},\dots,z_{i-1})$, $\alpha_i = f_{\alpha_i}(z_{1},\dots,z_{i-1})$, and $f_{\mu_i}$ and $f_{\alpha_i}$ are masked autoencoder neural networks (MADE~\cite{germain2015made}).
Let $\bm z$ be the input and $\hat{\bm z}$ be the output of a MADE network having $L$ hidden layers with $d_l$ nodes per layer, for $l=1,\cdots, L$. The mappings between the input and the first hidden layer, among hidden layers, and from the last hidden layer to the output are, respectively
\[
\left\{ \begin{matrix*}[l]
\bm{y}_{1}\!\!\!\!& = h_1(b_1 + (\bm W^1 \odot \bm M^1)\bm z), & \text{where } \bm M^1 \text{ is }d_1\times d \text{ and }\bm M^1_{u, v} = \mathbbm{1}_{m^{1}(u) \geq v}\\[0.7em]
\bm{y}_{l}\!\!\!\!& = h_l(b_l + (\bm W^l \odot \bm M^l)\bm{y}_{l-1}), & \text{where } \bm M^l \text{ is }d_{l}\times d_{l-1} \text{ and }\bm M^l_{u, v} = \mathbbm{1}_{m^{l}(u)\geq m^{l-1}(v)}\\[0.7em]
\hat{\bm{z}}\!\!\!\! & = h_{L+1}(b_{L+1} + (\bm W^{L+1} \odot \bm M^{L+1})\bm{y}_{L}) & \text{where } \bm M^{L+1} \text{ is }d \times d_{L} \text{ and }\bm M^{L+1}_{u, v} = \mathbbm{1}_{u > m^L(v)},\\
\end{matrix*}\right.
\]
where $h_l$ is the activation function between layer $l-1$ and $l$ for $l=1,\cdots, L+1$, $m^l(k)$ is a pre-set or random integer from $1$ to $d-1$, $b_l$ and $\bm W^l$ are the bias and weight parameters for $l=1,\dots,L+1$.

The matrix $\prod_{l=L+1}^{1}\bm M^{l}$, encoding the dependence between the components of the input $\bm z$ and output $\hat{\bm z}$, is strictly lower diagonal, thus satisfying the sought autoregressive property~\cite{germain2015made}. 
The masks $\bm M^{1},\dots,\bm M^{L+1}$ enable the computation of all $\mu_i$ and $\alpha_i$ in a single forward pass~\cite{papamakarios2017masked}; additionally, the Jacobian of each MAF layer is lower-triangular, with determinant equal to $|\det \partial f / \partial \bm z |^{-1} = \exp(\sum_{i=1}^{d} \alpha_i)$.

Between consecutive autoregressive layers and after the basic distribution,  batch normalization~\cite{ioffe2015batch} can be used to normalize the outputs from the previous layer so they have approximately zero mean and unit variance~\cite{papamakarios2017masked}. Batch normalization accelerates training by reducing the oscillations in the magnitudes of the flow parameters between consecutive layers, without complicating the computation of the Jacobian determinant in Eqn.~\eqref{equ:klDiv}. 
A batch normalization layer with input $\bm z$ and output $\hat{\bm z}$, parameterized through $\bm \beta$ and $\bm \gamma$, is
\[
\hat{\bm z} = F^{B}(\bm z) =\bm\beta\!+\!(\bm z - \bm m) \odot (\bm v + \epsilon)^{-1/2} \odot e^{\bm\gamma}\,\,\text{and}\,\, \left|\det \frac{\partial F^B}{\partial \bm z}\right| = \exp\left[\sum_i\left(\gamma_i - \frac{1}{2}\log(v_i + \epsilon)\right)\right],
\]
where $\bm m$ and $\bm v$ refer to the sample mean and variance of $\bm z$, respectively, and the tolerance $\epsilon$ (e.g. $10^{-5}$) ensures numerical stability if $\bm v$ has near zero components.

Besides MAF, RealNVP is another commonly used auto-regressive flow of the form
\[
\begin{cases}
\hat{\bm z}_{i} = \bm z_{i} & \text{for}\,\,i\le d'\\
\hat{\bm z}_{i} = \bm z_{i}\odot e^{\bm\alpha} + \bm \mu & \text{for}\,\, d' < i \le d
\end{cases},
\]
where $\bm \mu = f_\mu(z_{1},\dots,z_{d'})$ and $\bm \alpha = f_\alpha(z_{d'+1},\dots,z_{d})$. The output $\hat{\bm z}$ consists of identical copies of the input $\bm z$ for the first $d'<d$ elements, while the remaining $d-d'$ components are transformed by the MADE autoencoders $f_\mu$ and $f_\alpha$. 
MAF could be seen as a generalization of RealNVP by setting $\mu_i=\alpha_i=0$ for $i\leq d'$~\cite{papamakarios2017masked}.  We employ both RealNVP and MAF in our experiments in Section~\ref{sec:exp}.

\subsection{Surrogate Likelihood for Computationally Expensive Models}\label{sec:surrogate}

\noindent Consider a black-box model as a generic map $f: Z \to X$ between the random inputs $\bm{z} = (z_1, z_2, \cdots, z_d)^T \in Z$ and the outputs $(x_1, x_2,\cdots,x_m)^T \in X$, and assume $n$ observations $\bm x = \{\bm x_i\}_{i=1}^n \subset X$ to be available. Without loss of generality, we assume $\bm x$ to come from a Gaussian distribution with mutually independent components (other distributional assumptions can be made, depending on the specific problem).
With the Gaussian distribution assumptions, the log-likelihood of $\bm z$ given $\bm x$ is
\[
\ell(\bm z; f, \bm{x}) = -\frac{1}{2}\sum_{i=1}^n\sum_{j = 1}^m\left(\frac{f(\bm z)_j - x_{ij}}{\sigma_j}\right)^2 - \frac{n\cdot m}{2}\log(2\pi) - n\sum_{j=1}^m\log(\sigma_j).
\]

Our goal is to infer $\bm z$ and to quantify its uncertainty given $\mathbf{x}$. 
We employ a variational Bayesian paradigm and sample from the posterior distribution $p(\bm z\vert \bm x)\propto \ell(\bm z; f,\bm x) \pi(\bm z)$, with prior $\pi(\bm z)$ via NF. 
VI-NF requires the evaluation of the gradient of the cost function in Eqn.~\eqref{equ:klDiv} with respect to the NF parameters $\lambda$, replacing $p(\bm x, \bm z_K)$ with $p(\bm x\vert\bm z_K)\pi(\bm z_k)$ $=\ell(\bm z_K; f,\bm x) \pi(\bm z_k)$, and approximating the expectations with their MC estimates. 
However, the likelihood function needs to be evaluated at every MC realization, which can be costly if the model $f(\bm{z})$ is computationally expensive. In addition, automatic differentiation through a legacy (e.g. physics-based) solver may be an impractical, time-consuming, or require the development of an adjoint solver.

One solution is to replace the model $f$ with a computationally inexpensive surrogate $\hat{f}: Z \times \Omega \to X$ parameterized by $\omega \in \Omega$, whose derivatives can be obtained at a relatively low computational cost, but intrinsic bias in the selected surrogate formulation (i.e., $f(\bm{z}) \notin \{\hat f (\bm{z}; \omega)|\forall \omega \in \Omega\}$), a limited number of training examples, and locally optimal $\omega$ can compromise the accuracy of $\hat{f}$.

To resolve these issues, we propose to update the surrogate model adaptively by smartly weighting the samples of $\mathbf{z}$ from NF. 
Once a newly updated surrogate is obtained, the likelihood function is updated, leading to a new posterior distribution that will be approximated by VI-NF, producing, in turn, new samples for the next surrogate model update, and so on. In the next section, we will introduce this new approach in detail and provide its algorithm.

\subsection{Variational Inference via Normalizing Flow with an Adaptive Surrogate (NoFAS)}\label{sec:ourAlgorithm}

\noindent The computational cost of training a surrogate model with an uniform accuracy over the entire parameter space $Z$ grows exponentially with the problem dimensionality (see, e.g.,~\cite{montgomery2017design}). 
In such a case, many training samples would correspond to model outputs that are relatively far from the available observations $\mathbf{x}$ and contribute minimally to learning the posterior distribution of $\mathbf{z}$ given $\mathbf{x}$, leading to a massive waste of computational resources.

\begin{figure}[!htb]
\centering
\begin{tikzpicture}[node distance = 2cm, auto]
    \node [block]                                       (obs)   {Observation $\bm{x}$};
    \node [block, right of=obs, node distance=3cm]      (like)  {Likelihood $\ell(\bm z; \hat{f}, \bm{x})$};
    \node [block, right of=like, node distance=3cm]     (sur)   {NN Surrogate $\hat{f}$};
    \node [block, right of=sur, node distance=4cm]      (model) {Model $f$};
    \node [block, below of=obs]                         (prior) {Prior $\pi(\bm z)$};
    \node [block, below of=prior]                       (noise) {Basic Dist. $q_0(\bm z)$};
    \node [block, right of=prior, node distance=3cm]    (post)  {Posterior $p(\bm z|\hat{f}, \bm{x})$};
    \node [block, right of=noise, node distance=3cm]    (nf)    {NF $F(\bm z;\lambda)$};
    \node [block, right of=nf, node distance=3cm]       (appr)  {Approx. posterior $q_K(\bm z_K)$};
    \node [block, below of=model]                       (param) {Sample $Z_G$ from NF};
    \node [block, right of=model, node distance=3cm]    (pre)   {Pre-grid $Z_P$};
    \node [draw,inner xsep=4mm,inner ysep=7mm,fit=(model)(pre)(param),label={}] (big){};  
    \path [line]    (obs)   --  (like);
    \path [line]    (sur)   --  (like);
    \draw [line]    (big.west) -- ++(0,1) --  node[above]{train}(sur);
    \path [line]    (prior) --  (post);
    \path [line]    (post)  --  node [right]{train}(nf);
    \path [line]    (like)  --  (post);
    \path [line]    (noise) --  (nf);
    \path [line]    (nf)    --  (appr);
    \path [line]    (appr)  -|  node [below]{calibrate}(param);
    \path [line]    (param) --  (model);
    \path [line]    (pre)   --  (model);
    \path [line]    (appr)  |-  (post);
\end{tikzpicture}
\caption{Illustrative Diagram for the NoFAS algorithm.}\label{fig:asnf_scheme}
\vspace{-12pt}
\end{figure}
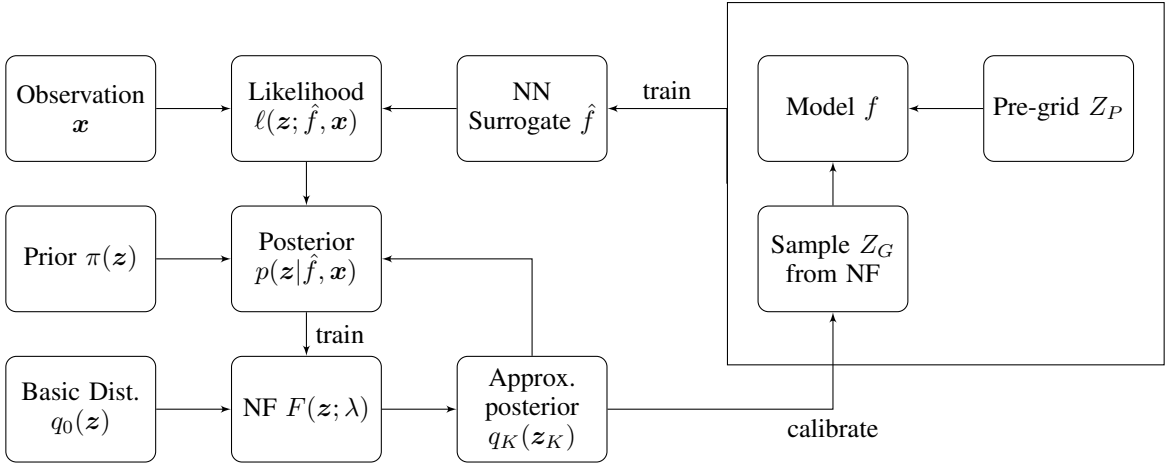

Motivated by this observation, we propose a strategy to alternate gradient-based updates for the surrogate model parameters $\omega$ and NF parameters $\lambda$. 
Once every $c$ normalizing flow iterations (\emph{calibration frequency}), a sub-sample from the \emph{batch} $\{\bm{z}_{K}^{(s)}\}_{s=1}^{b}$ (used in Eqn.~\eqref{equ:klDiv} to evaluate the MC expectations) is used to provide additional training samples to \emph{adaptively} improve the surrogate $\hat{f}$.
This approach, referred to as VI using Normalizing Flow with an Adaptive Surrogate (NoFAS for short), is illustrated in Figure~\ref{fig:asnf_scheme} and its algorithmic steps are presented in Algorithm~\ref{alg:asnf}.

NoFAS starts with an initial surrogate model trained from an a-priori selected \emph{pre-grid}, $Z_P = \{\bm{z}_P^{(s)}\}_{s=1}^{S_P}$ of realizations from $Z$ (typically tensor product grids or low-discrepancy sequences~\cite{dick2010digital}).
At every $(j\cdot c),\,j\in\mathbb{N}$ flow parameter update, $S_G < b$ realizations $Z_{G,j} = \{\bm{z}^{(t)}_{G,j}\}_{t=1}^{S_{G}}$ are sub-sampled at random from the batch $\{\bm{z}_{K}^{(s)}\}_{s=1}^{b}$. 
The solutions from the true model $f(Z_{G, j})$ are computed and used to update the surrogate model $\hat{f}$ on the training set
\begin{equation}
Z_{T} = \left(\cup_{\alpha=\max(1, j-M+1)}^j Z_{G, \alpha}\right) \cup Z_P. 
\end{equation}

Additionally, the sequentially collected samples carry different weights in the loss function for the surrogate model update; more recently collected parameter samples $Z_{G}$ receive a larger weight than those collected earlier so to achieve better accuracy around regions of high posterior density progressively discovered by NF. Larger weights are also assigned to the pre-grid realizations $Z_{P}$, since they are responsible to ensure some \emph{generalizability} of the trained surrogate over the parameter space $Z$, rather than just capturing local features. Taken together, if a $l_2$ norm is used (but other types of loss can be chosen), the surrogate model optimization loss would take the form
\begin{equation}\label{equ:surrogate_loss}
\begin{matrix*}[l]
\ds L_j(\beta_{0},\beta_{1},\omega,Z_{P},Z_{G}) =  \beta_0\sum_{s=1}^{S_P} \|\hat{f}(\bm{z}^{(s)}_P; \omega) - f(\bm{z}^{(s)}_P)\|_2^2 / S_P + \\
\ds+ (1-\beta_0)\sum_{\alpha = \max(1, j-M+1)}^j\sum_{s=1}^{S_G} \sigma(\exp(-\beta_1(j-\alpha))\,\|\hat{f}(\bm{z}^{(s)}_{G, \alpha}; \omega) - f(\bm{z}^{(s)}_{G, \alpha})\|_2^2 / S_G,
\end{matrix*}
\end{equation}
where $M$ denotes the \emph{memory} of the proposed adaptive scheme, i.e., the number of the more recent realizations in $Z_{G}$ included in the loss, $\beta_0$ represents the weight assigned to the pre-grid realizations, $\beta_1 > 0$ defines the rate of exponentially decaying weights for $Z_{G, \alpha},\,\alpha=\max(1, j-M+1),\dots,j$ (with realizations ordered from the most recent $j$ to the least recent $\max(1, j-M+1)$) and $\sigma(\cdot)$ is the softmax function.

In our numerical experiments, the behaviors of batch $\{\bm z^{(s)}_{K}\}_{s=1}^{b}$ could be observed to evolve in three phases.
In the first phase, the samples are aggregated in clusters occupying a small region of the parameter space. 
In the second phase, the aggregated clusters move together to a region of high posterior density, followed by a third stage where they are scattered to better cover the posterior distribution.
Batch sub-samples extracted in the first two phases can be similar in some cases and lack diversity to provide a good characterization of the local response of the true model $f$, negatively affecting convergence. Additionally, evaluating the true model at almost identical inputs (parameter samples) is a waste of computational resources.
A possible remedy is to perturb the parameter realizations by injecting Gaussian noise, before storing them in $Z_{G,j}$. For the experiments presented in Section~\ref{sec:exp}, we added Gaussian noise $\mathcal{N}(0, \varepsilon^2)$ if the batch has a standard deviation $<\varepsilon$ and we used $\varepsilon=0.1$. We conjecture that several factors affect the choice of $\varepsilon$, such as the prior knowledge about the parameters and the local curvature of the true model response; future work is needed to better understand this phenomenon and to develop a systematic approach for choosing $\varepsilon$.
\begin{algorithm}[H]\label{alg:asnf}
\SetKwInOut{Input}{input}\SetKwInOut{Output}{output}
\Input{Model $f$, observations $\bm{x}$, batch size $b$, calibration frequency $c$ and size $S_G$}
\BlankLine
Generate initial surrogate model training set $Z_P = \{\bm{z}_P^{(s)}\}_{s=1}^{S_0}$ from prior $\pi(\bm{z})$\;

Let $\omega_0 = \arg\min_{\omega\in\Omega}L(\omega; Z_P)$ and initialize surrogate $\hat{f}(\cdot \ ; \omega_0)$\;
Initialize$^{\dagger}$ flow parameters $\lambda_0$ and set $t=0$\; 
Initialize hyper-parameters (e.g. learning rate $\eta_0$, learning scheduler) for optimizer $\psi_{F}(\lambda, \mathcal{F}(\bm x))$ over $\lambda$ that minimizes the free energy bound in Eq.~\eqref{equ:klDiv}\;

\While{$t \leq T_{F}$ or $|\mathcal{F}_t(\bm x) - \mathcal{F}_{t-1}(\bm x)| > \epsilon$}{

    Obtain a batch of samples $\{\bm{z}^{(s)}_0\}_{s=1}^{b}$ from a basic distribution (e.g., $\N(\bm{0}, \bm{I}_d)$)\;
    
    Compute  $\bm{z}^{(s)}_{K} = F(\bm{z}^{(s)}_{0}; \lambda_t),\,s=1,\dots,b$\; 
    
    \If{$(t \text{ mod } c == 0)$}{

        Randomly draw a subset of $\{\bm{z}^{(s)}_K\}_{s=1}^{b}$ and push it into $Z_{G,t}$\;
        
        Reset the parameters of the scheduler$^{\ddagger}$ for optimizer $\psi_{S}(\omega, L_t(\beta_{0},\beta_{1},\omega,Z_{P},Z_{G}))$ over $\omega$, minimizing the surrogate loss in Eq.~\eqref{equ:surrogate_loss}\;
        
        \For{$\tau = 0,1,\cdots, T_{S}$}{
            $\omega_{\tau+1}\leftarrow\psi_S^{(t)}(\omega_{\tau}, L_t(\beta_{0},\beta_{1},\omega,Z_{P},Z_{G}))$\;
        }
    }
    Update the likelihood $\ell(\bm z_K^{(s)};\hat{f}(\cdot \ ; \omega_{T_{S}}), \bm{x})$ for $s = 1,\cdots, b$\;
    $\lambda_{t+1}\leftarrow\psi_{F}(\lambda_t, \mathcal{F}(\bm x))$\;
    $t \leftarrow t + 1$\;
}
\caption{NoFAS algorithm.}
\end{algorithm}
\noindent{\small$^{\dagger}$ We used the uniform Glorot initialization~\cite{glorot2010understanding} in the experiments in Section \ref{sec:exp}. However, other initialization schemes, such as Kaiming Uniform~\cite{he2015delving} can also be used.}\\
\noindent{\small$^{\ddagger}$The learning rate scheduler parameter reduces the learning rate at every iteration. Our algorithm \emph{resets} the learning rate back to $\eta_0$  at every surrogate model update.}

\section{Experiments}\label{sec:exp}

\noindent In this section, we run four numerical experiments to demonstrate the application and performance of NoFAS in parameter estimation, for models formulated as algebraic or differential equations. The first two experiments have identifiable model parameters, the model in the third experiment has highly correlated parameters, and we purposely design an over-parameterized model in the fourth experiment to examine the robustness of the NoFAS procedure for variational inference. In each of the four experiments, we employed a fully connected neural network with two hidden layers with $64$ and $32$ nodes, respectively, as the surrogate model, and set $\beta_0 = 0.5$ and $\beta_1 = 0.1$. We investigated numerically the sensitivity of NoFAS to different choices of $\beta_0$ and $\beta_1$ and the results are presented in Appendix D.1. The results suggest that NoFAS performs well for $\beta_0 \in [0.5, 0.7]$ and $\beta_1 \in [0.01, 1.0]$.
We also compare NoFAS with MH, BBVI, and NF with a fixed surrogate model, evaluating their performance based on the accuracy in the recovered posterior and predictive posterior distributions, and the computational cost savings as measured by the number of true model evaluations. 

\subsection{Experiment 1: Model with Closed-form Solution}\label{sec:2DSimple}

\noindent The output from model $f:\mathbb{R}^{2}\to \mathbb{R}^{2}$ in this experiment has the closed-form expression
\begin{equation}
f(\bm z) = f(z_{1},z_{2}) = (z_1^3 / 10 + \exp(z_2 / 3), z_1^3 / 10 - \exp(z_2 / 3))^T. 
\end{equation}
Observations $\widetilde{\bm{x}}$ are generated as
\begin{equation}\label{eqn:exp1}
\widetilde{\bm{x}} = \bm{x}^{*} + 0.05\,|\bm{x}^{*}|\,\odot\bm{x}_{0},
\end{equation}
where $\bm{x}_{0} \sim \N(0,\bm I_2)$. 
We set the \emph{true} model parameters at $\bm{z}^{*} = (3, 5)^T$, with output $x^{*} = f(\bm z^{*})=(7.99, -2.59)$, and simulated 50 sets of observations from Eqn.~\eqref{eqn:exp1}. The likelihood of $\bm z$ given the observed data  $\widetilde{\bm{x}}$ is Gaussian and we adopt a noninformative uniform prior $\pi(\bm z)$.

For surrogate model estimation, we set the maximum number of true model evaluations -- referred to as the \emph{budget} -- at $64$, and examine two scenarios for allocating the budget. In the first scenario, the entire budget is assigned to a 2-dimensional $8\times 8 = 64$ pre-grid $Z_p$. The surrogate $\hat{f}$ is learned from $Z_{P}$ only, and never updated.
In the second scenario, we allocate a budget of $4\times4=16$ model solutions to $Z_{P}$ and use the rest to calibrate $\hat{f}$ with $S_G=2$ samples every $c=1000$ NF iterations. 
We refer to these two scenarios as the \emph{fixed} and \emph{adaptive} surrogate, respectively, where the latter coincides with NoFAS.
A RealNVP architecture is employed, alternating 5 batch normalization layers and 5 linear masked coupling layers having 1 hidden layer of 100 neurons. We run the RMSprop optimizer with a learning rate of $0.03$ and an exponential decay coefficient of $0.9995$ to update $\lambda$.

\begin{figure}[!htb]
\centering
\includegraphics[width=1.0\textwidth]{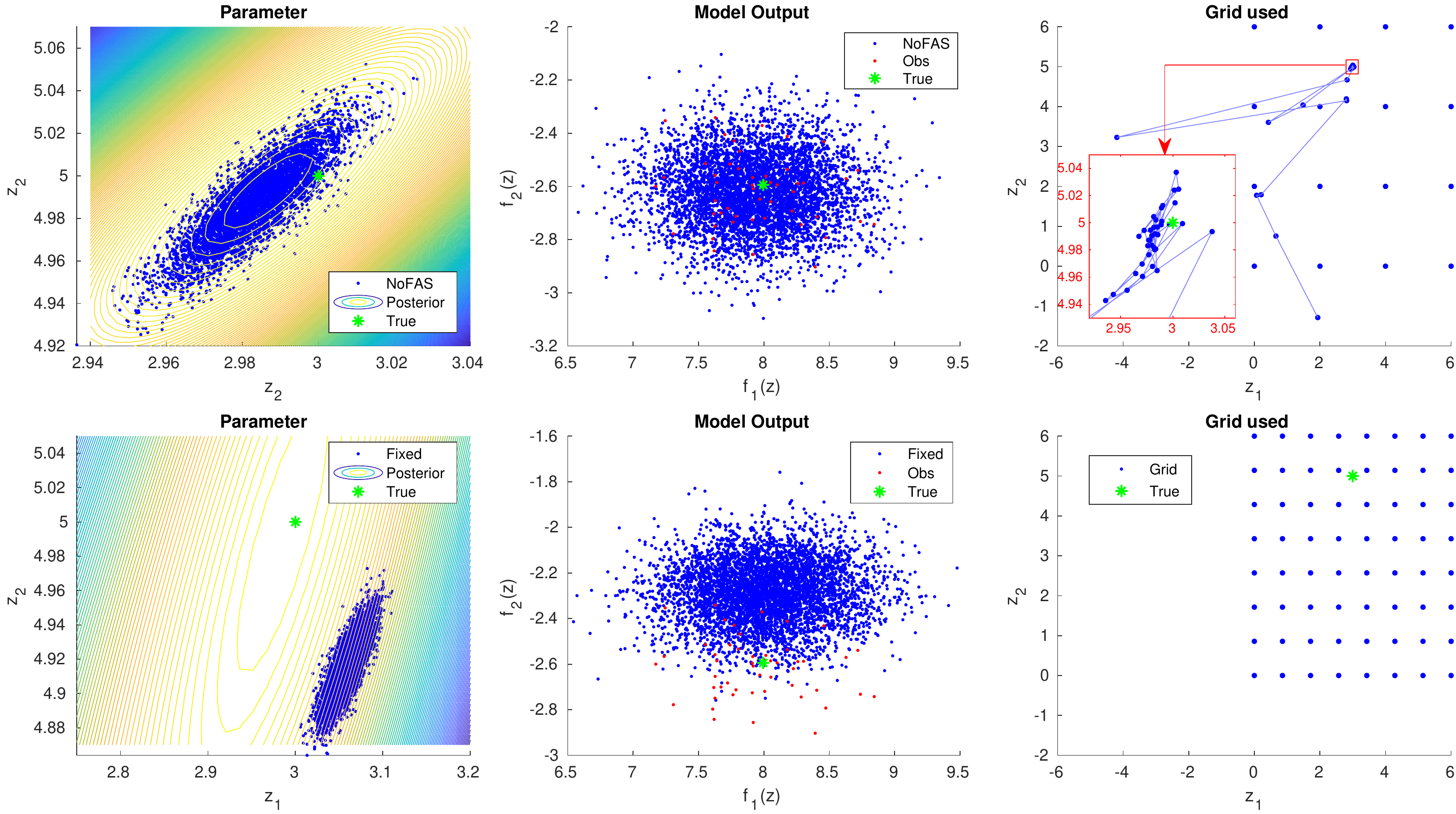}
\caption{Parameter samples from the posterior distribution and the corresponding model solutions using NoFAS (first row) and a fixed surrogate (second row) in Experiment 1. The plots in the first column show the posterior parameter realizations (blue points), the true parameter values $\bm z^{*}$ (green star), and the contour for the true posterior distribution of $\bm z$. The plots in the second column display the observations $\widetilde{\bm x}$ (red dots), the posterior predictive samples, and $\bm x^{*}$ (green star). The pre-grid and the adaptive realizations in $Z_{G}$ are illustrated in the third column.}\label{fig:ex1}
\end{figure}

The results are presented Figure~\ref{fig:ex1}. The adaptive surrogate is less biased and leads to an improved quantification of uncertainty compared to the fixed surrogate. 
By better capturing $f$ locally around $\bm z^{*}$ (third column in Figure~\ref{fig:ex1}), the adaptive surrogate also generates a posterior predictive distribution that agrees with the observations (second column in Figure~\ref{fig:ex1}).

\begin{figure}[!htb]
\centering
\includegraphics[width=0.7\textwidth]{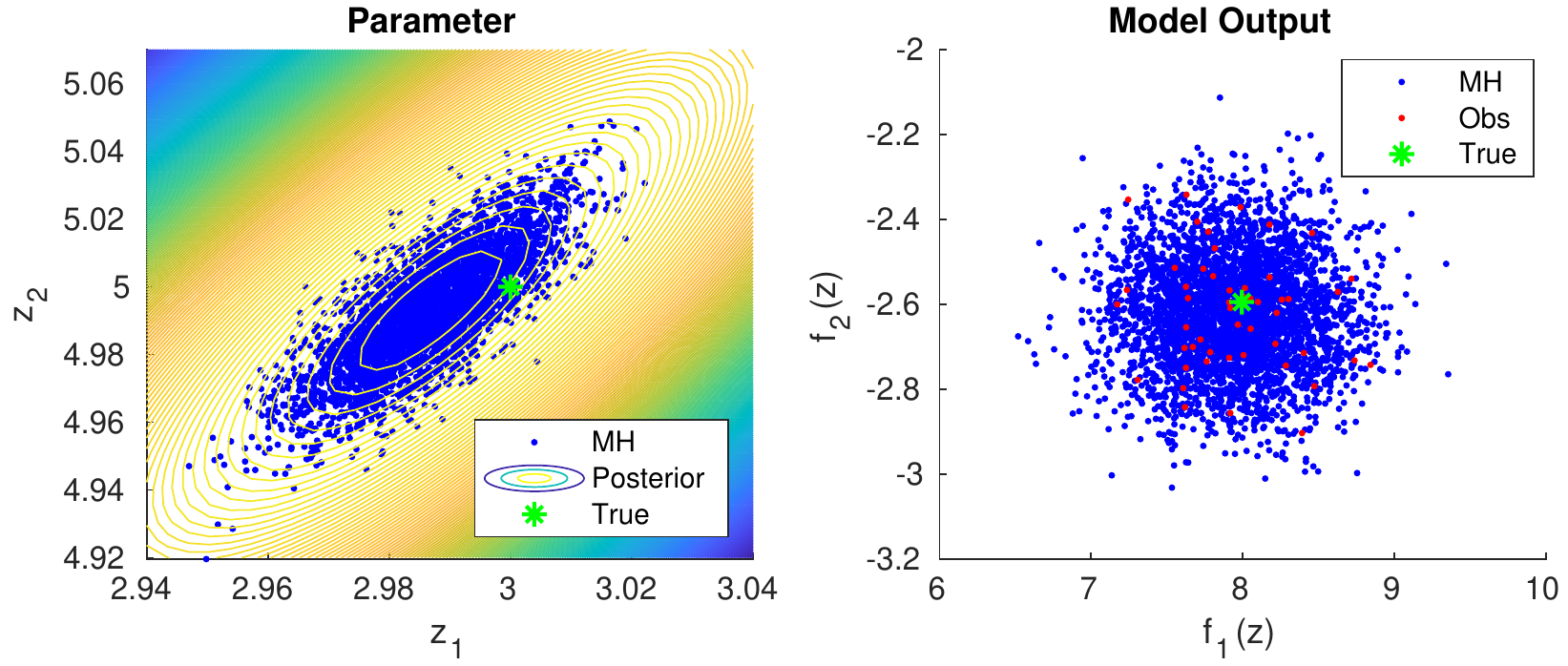}
\caption{Parameters and model solutions from Experiment 1 using a MH sampler. The blue points represent the parameter samples of $\bm z$ from MH (left plot), the corresponding model solutions $\bm x$ (right plot). A green star represents the true parameters $\bm z^{*}$ (left) and model solutions $\bm x^{*}$ (right). Red dots indicate the observations $\widetilde{\bm x}$. A contour plot of the true posterior distribution is also superimposed to the posterior samples on the left plot.}\label{fig:trivial_MH}
\end{figure}

We also run the MH algorithm to obtain the posterior samples on $\bm z$ and present the results in Figure~\ref{fig:trivial_MH}.
MH used $4\times 10^6$ iterations, resulting in $3600$ effective samples using a burn-in and thinning rate of $10\%$ and $1/1000$, respectively.
The accuracy of MH in capturing the posterior and posterior predictive distributions is similar to NoFAS, but MH evaluated the true model $4\times 10^6$ times compared to only $64$ times for NoFAS.

\subsection{Experiments 2 and 3: ODE Hemodynamic Models}\label{sec:ode_models}

\noindent In this section, NoFAS is applied to estimate the parameters of a two-element and a three-element Windkessel models. These are lumped parameter (or circuit) models of the interaction between flow, pressure, resistance and compliance in a vascular system. These models are also frequently used to mimic physiological boundary conditions in numerical hemodynamics, and the solutions of inverse problems for Windkessel models is often a prerequisite to determine boundary conditions parameters so that simulation outputs can match patient-specific responses.
These systems also offer a concise representation (i.e., with a small number of parameters) of circulatory sub-systems in humans, and therefore constitute an ideal test bed for the application of advanced inference algorithms to physiological models with many more parameters.

\subsubsection{Experiments 2: Two-element Windkessel Model}

\begin{figure}[!ht]
\centering
\begin{subfigure}[c]{0.5\linewidth}
\centering
z\includegraphics[width=\linewidth]{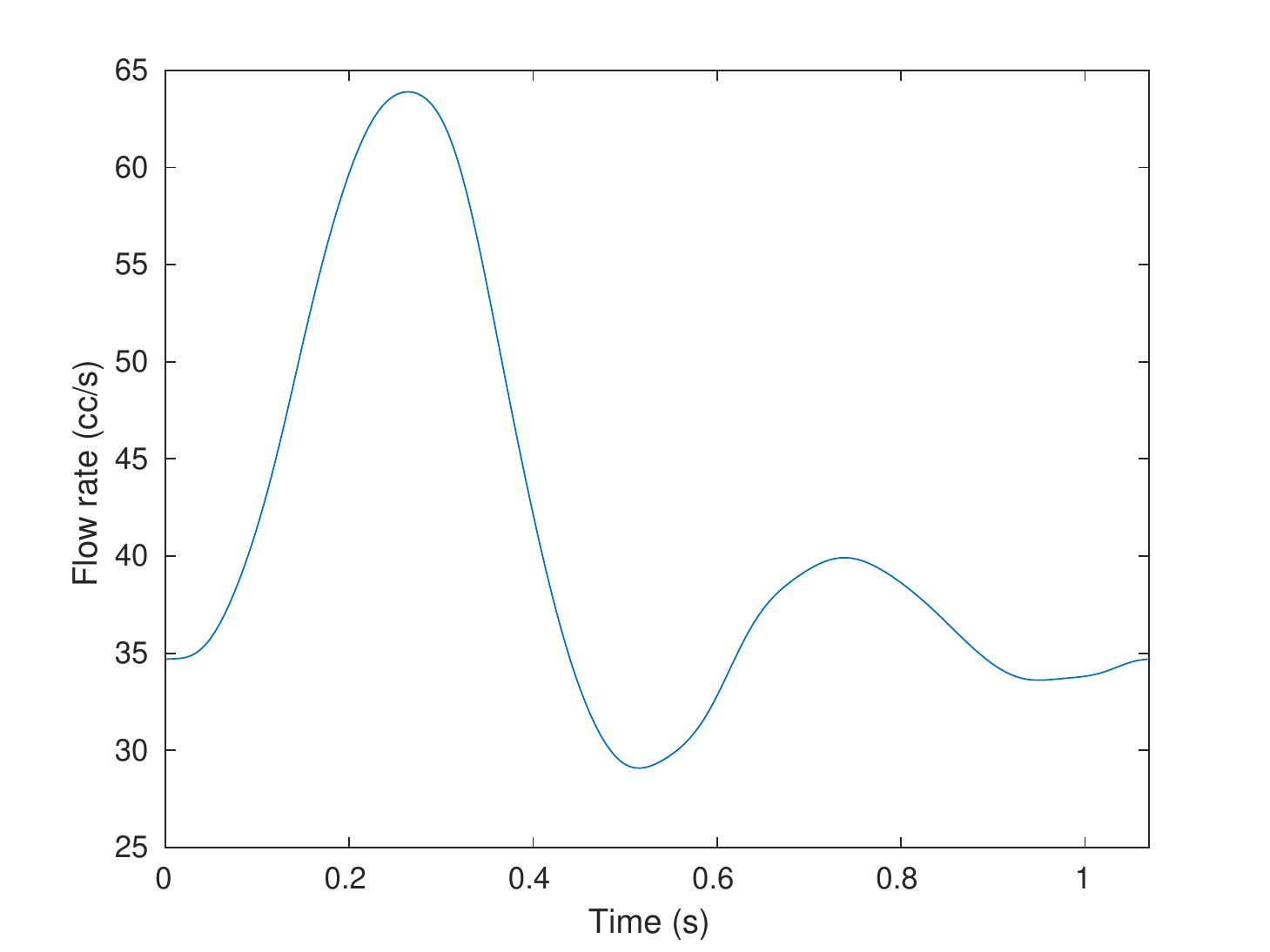}
\caption{Time history for proximal flow rate $Q_{p}$.}
\end{subfigure}
\begin{subfigure}[c]{0.45\linewidth}
\centering
\begin{subfigure}{\linewidth}
\centering
\includegraphics[width=0.8\linewidth]{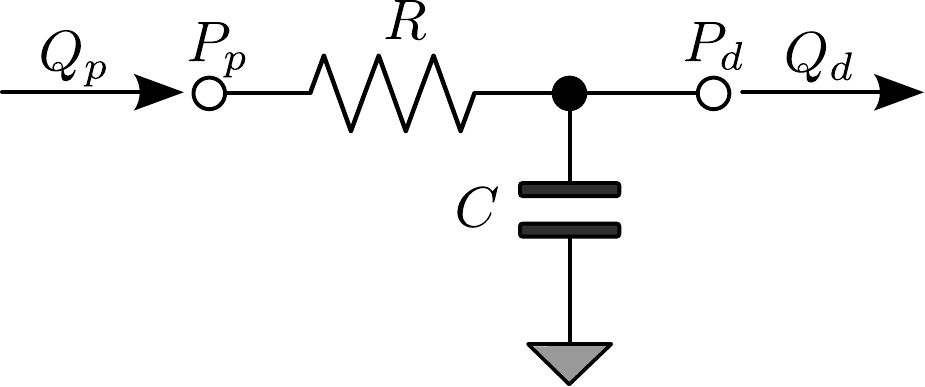}
\caption{Two-element Windkessel model.}
\end{subfigure}

\vspace{18pt}
\begin{subfigure}{\linewidth}
\centering
\includegraphics[width=\linewidth]{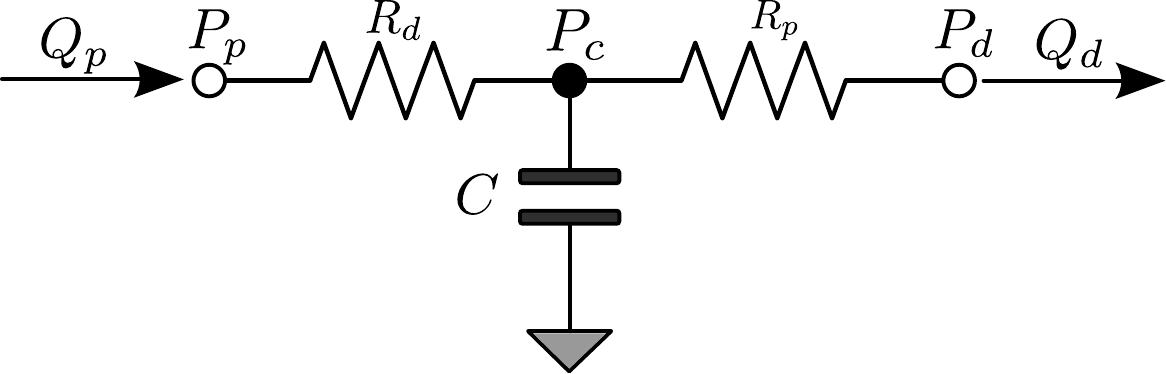}
\caption{Three-element Windkessel model.}
\end{subfigure}
\end{subfigure}
\caption{RC and RCR hemodynamic models.}\label{fig:circuitModels}
\end{figure}

\noindent The two-element Windkessel model was developed by Otto Frank~\cite{frank1990basic} and, when applied to the flow-pressure relation in the aorta, is successful in explaining the exponential decay of the aortic pressure following the closure of the aortic valve. 
This model requires two parameters, i.e., a systemic resistance $R \in [100, 1500]$ Barye$\cdot$ s/ml and a systemic capacitance $C \in [1\times 10^{-5}, 1 \times 10^{-2}]$ ml/Barye, which are responsible for its alternative name as \emph{RC model}. We provide a periodic time history of the aortic flow in Figure~\ref{fig:circuitModels} and use the RC model to predict the time history of the proximal pressure $P_{p}(t)$, specifically its maximum (max), minimum (min) and average (ave) values over a typical heart cycle, while assuming the distal resistance $P_{d}(t)$ as a constant in time, equal to 55 mmHg. In our experiment, we set the true resistance and capacitance as $z_{1}^{*}=R^{*} = 1000$ Barye$\cdot$ s/ml and $z_{2}^{*}=C^{*} = 5\times 10^{-5}$ ml/Barye and determine $P_{p}(t)$ from a RK4 numerical solution of the following algebraic-differential system of two equations
\begin{equation}\label{equ:RC}
Q_{d} = \frac{P_{p}-P_{d}}{R},\quad \frac{\df P_{p}}{\df t} = \frac{Q_{p} - Q_{d}}{C},
\end{equation}
where $Q_{p}$ is the flow entering the RC system and $Q_{d}$ is the distal flow (see Figure~\ref{fig:circuitModels}).
Synthetic observations are generated by adding Gaussian noise to the true model solution $\bm{x}^{*}=(P_{p,\text{min}},$ $P_{p,\text{max}},$ $P_{p,\text{ave}})= (78.28, 101.12,  85.75)$, i.e., $\widetilde{\bm{x}}$ follows a multivariate Gaussian distribution with mean $\bm{x}^{*}$ and a diagonal covariance matrix with entries $0.05\,x_{i}^{*}$, where $i=1,2,3$ corresponds to the maximum, minimum, and average pressures, respectively. The aim is to quantify the uncertainty in the RC model parameters given 50 repeated pressure measurements. We imposed a non-informative prior on $R$ and $C$. Also note that the two-element Windkessel model is \emph{identifiable}. The resistance $R$ can be estimated based on the mean flow and pressure, while the capacitance $C$ directly affects the pulse pressure (i.e. the difference between systolic and diastolic pressure).

Similar to Experiment 1, we consider a total budget of $64$ output evaluations via the true model $f$ and compare NoFAS with the fixed surrogate approach, a MH sampler, and BBVI. For NoFAS, $M=20$ batches of $S_{G}=2$ samples per batch are collected at a calibration frequency of $c=1000$ NF parameter updates. The NF architecture is MAF composed of five alternated batch normalization and 5 MADE layers, each consisting of a MADE autoencoder with 1 hidden layer of 100 nodes and ReLU activation.

\begin{figure}[!htb]
\centering
\includegraphics[width=1.0\textwidth]{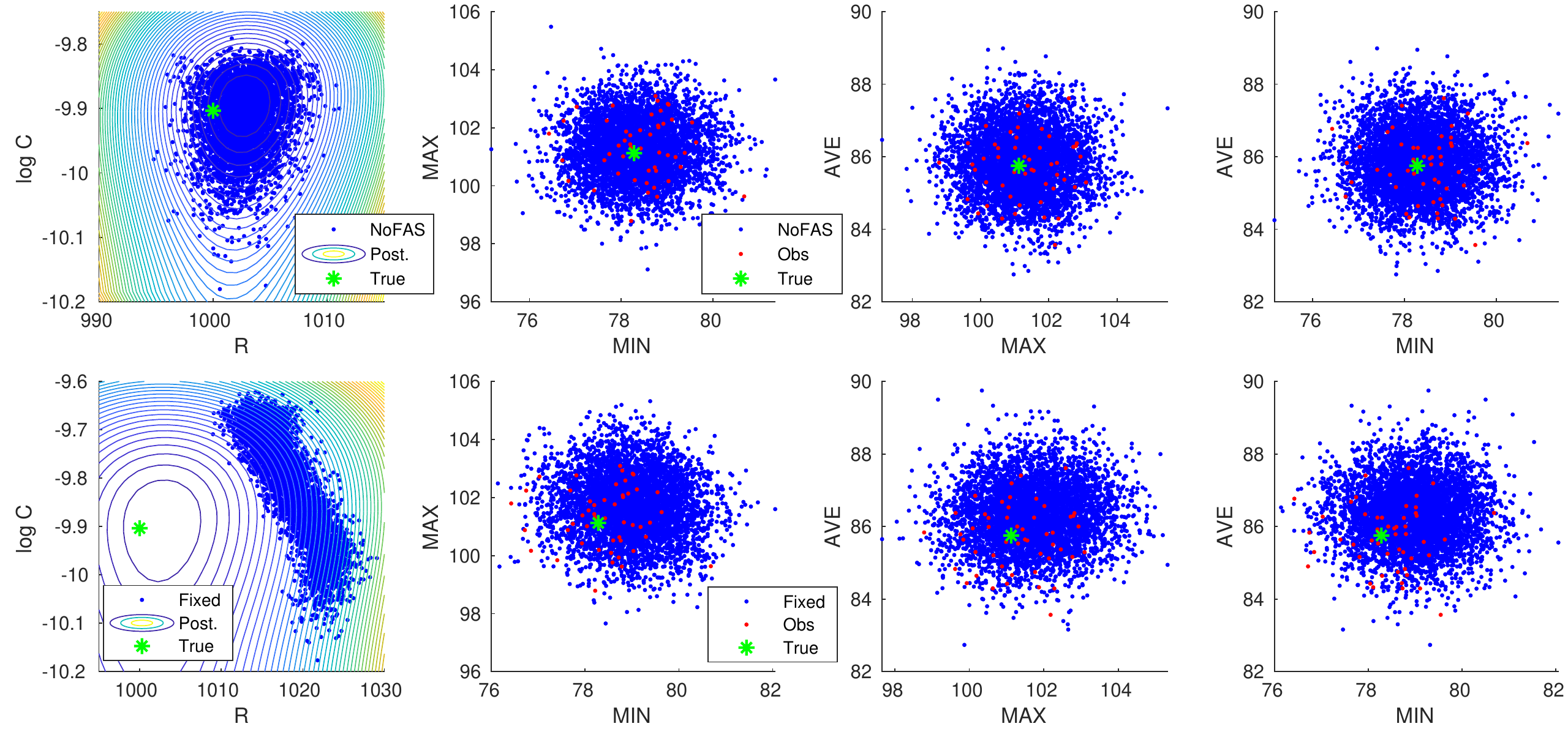}
\caption{Parameters and model solutions from inference using NoFAS (first row) and NF with a fixed surrogate (second row) in Experiment 2. The first column shows the posterior samples $\bm{z}_{K}$ while the other columns illustrate the corresponding model outputs. Symbols and colors are consistent with those in Figure~\ref{fig:ex1}. }\label{fig:RC}
\end{figure}

The results from NoFAS and from NF with a fixed surrogate are presented in Figure~\ref{fig:RC}. The approximate posterior distribution of the RC model parameters with a fixed surrogate is clearly biased, and so is the corresponding predictive posterior distribution. In contrast, NoFAS provides significantly more accurate results. 

\begin{figure}[!htb]
\centering
\includegraphics[width=1.0\textwidth]{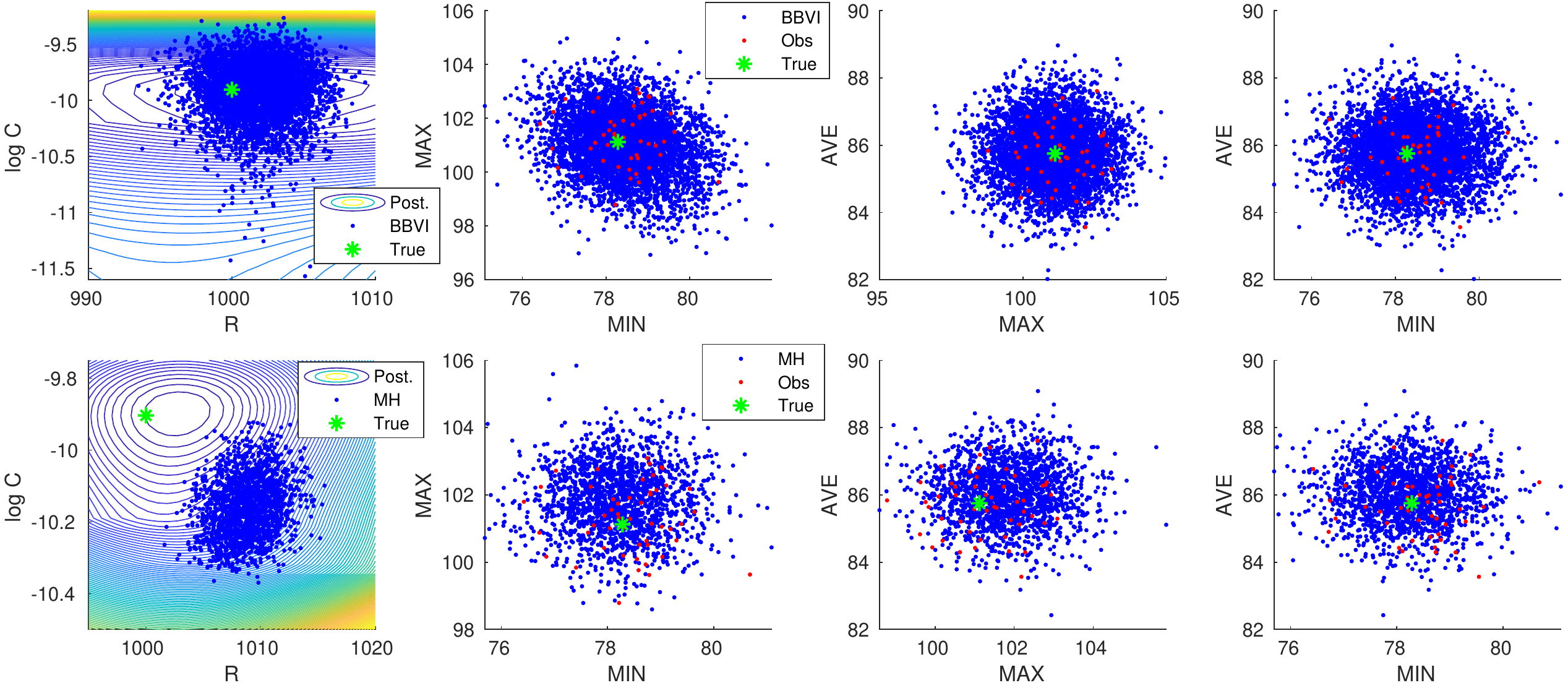}
\caption{Parameters and model solutions from BBVI (first row) and MH (second row) for Experiment 2. Symbols and colors are consistent with those in Figure~\ref{fig:ex1}.}\label{fig:RC_BBVI_MH}
\end{figure}

The results from BBVI and MH are presented in Figure~\ref{fig:RC_BBVI_MH}. BBVI uses a normal variational distribution. For the RC model, where no posterior correlation between parameters $R$ and $C$ is expected, BBVI with the mean field assumption captures the posterior and predictive posterior distributions accurately. MH operates on a fixed surrogate trained using a $30\times30$ pre-grid and requires $2\times10^6$ true model evaluations. A total of 1800 effective posterior samples were generated using a burn-in and thinning rate of $10\%$ and $1/1000$, respectively. Even though this fixed surrogate is trained from a large number of examples, it still introduces bias in the estimated parameters.
The posterior samples from MH deviate instead significantly from the true model parameters, and so do the posterior predictive samples, though less pronounced.

\subsubsection{Experiments 3: Three-element Windkessel Model}\label{sec:RCR}

\noindent The three-parameter Windkessel or \emph{RCR} model is characterized by proximal and distal resistance parameters $R_{p}, R_{d} \in [100, 1500]$ Barye$\cdot$ s/ml and one capacitance parameter $C \in [1\times 10^{-5}, 1\times 10^{-2}]$ ml/Barye.  
Even if it consists of a relatively simple lumped parameter formulation, the RCR circuit model is not identifiable. The average distal pressure is only affected by the total system resistance, i.e. the sum $R_{p}+R_{d}$, leading to a negative correlation between these two parameters. Thus, an increment in the proximal resistance is compensated by a reduction in the distal resistance (so the average distal pressure remains the same) which, in turn, reduces the friction encountered by the flow exiting the capacitor. An increase in the value of $C$ is finally needed to restore the average, minimum and maximum pressure. This leads to a positive correlation between $C$ and $R_{d}$.

Similar to the RC model, the output consists of the proximal pressure $P_{p}(t)$, specifically its maximum, minimum and average values $(P_{p,\text{min}}, P_{p,\text{max}}, P_{p,\text{ave}})$ over one heart cycle.
The true parameters are $z^{*}_{1} = R^{*}_{p} = 1000$ Barye$\cdot$s/ml, $z^{*}_{2}=R^{*}_{d} = 1000$ Barye$\cdot$s/ml and $C^{*} = 5\times 10^{-5}$ ml/Barye and the proximal pressure is computed from the solution of the algebraic-differential system
\begin{equation}
Q_{p} = \frac{P_{p} - P_{c}}{R_{p}},\quad Q_{d} = \frac{P_{c}-P_{d}}{R_{d}},\quad \frac{\df P_{c}}{\df t} = \frac{Q_{p}-Q_{d}}{C},
\end{equation}
where the distal pressure is set to $P_{d}=55$ mmHg.
Synthetic observations are generated from $N(\bm\mu, \bm\Sigma)$, where $\mu=(f_{1}(\bm{z}^{*}),f_{2}(\bm{z}^{*}),f_{3}(\bm{z}^{*}))^T = (P_{p,\text{min}}, P_{p,\text{max}}, P_{p,\text{ave}})^T = (100.96,$ $148.02,$ $ 116.50)^T$ and $\bm\Sigma$ is a diagonal matrix with entries $(5.05, 7.40, 5.83)^T$. The budgeted number of true model solutions is $216$; the fixed surrogate model is evaluated on a $6\times 6\times 6 = 216$ pre-grid while the adaptive surrogate is evaluated with a pre-grid of size $4\times 4\times 4 = 64$ and the other 152 evaluations are adaptively selected. 
The NF architecture and hyper-parameter specifications are the same as for the RC model, except a more frequent surrogate update of $c = 300$ and a larger batch size $b = 500$.

\begin{figure}[!htb]
\centering
\includegraphics[width=1.0\textwidth]{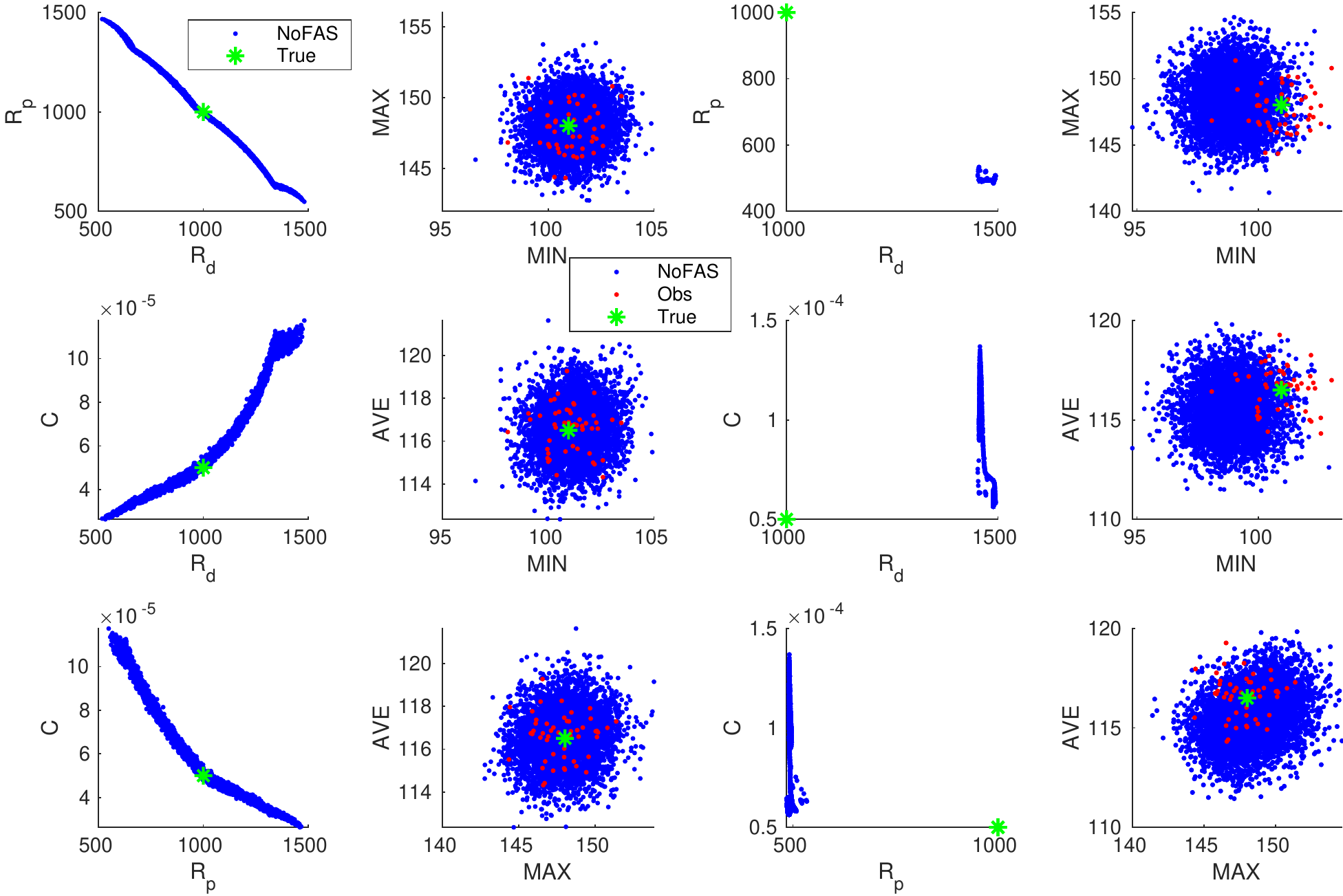}
\caption{Parameter posterior samples and corresponding model solutions for the RCR model. The two columns on the left show the results of NoFAS, while the two columns on the right contain the results for NF with a fixed surrogate. The first and third columns show the posterior samples, and the second and fourth columns show the posterior predictive samples (blue) and observations (red). The green stars represent ether the true parameters $R^{*}_{p}$, $R^{*}_{d}$, $C^{*}$ (1st and 3rd columns) or the corresponding model solutions (2nd and 4th columns).}\label{fig:exp2.1}
\end{figure}

The results are presented in Figure~\ref{fig:exp2.1}. The posterior samples obtained through NoFAS capture well the non-linear correlation among the parameters and generate a fairly accurate posterior predictive distribution that overlaps with the observations but has a slightly larger dispersion, as expected.
In contrast, NF with a fixed surrogate fails to capture the parameter correlations and the complex shape of the posterior distribution; in addition, the posterior predictive samples deviate significantly from the observed data.

\begin{figure}[!htb]
\centering
\includegraphics[width=1.0\textwidth]{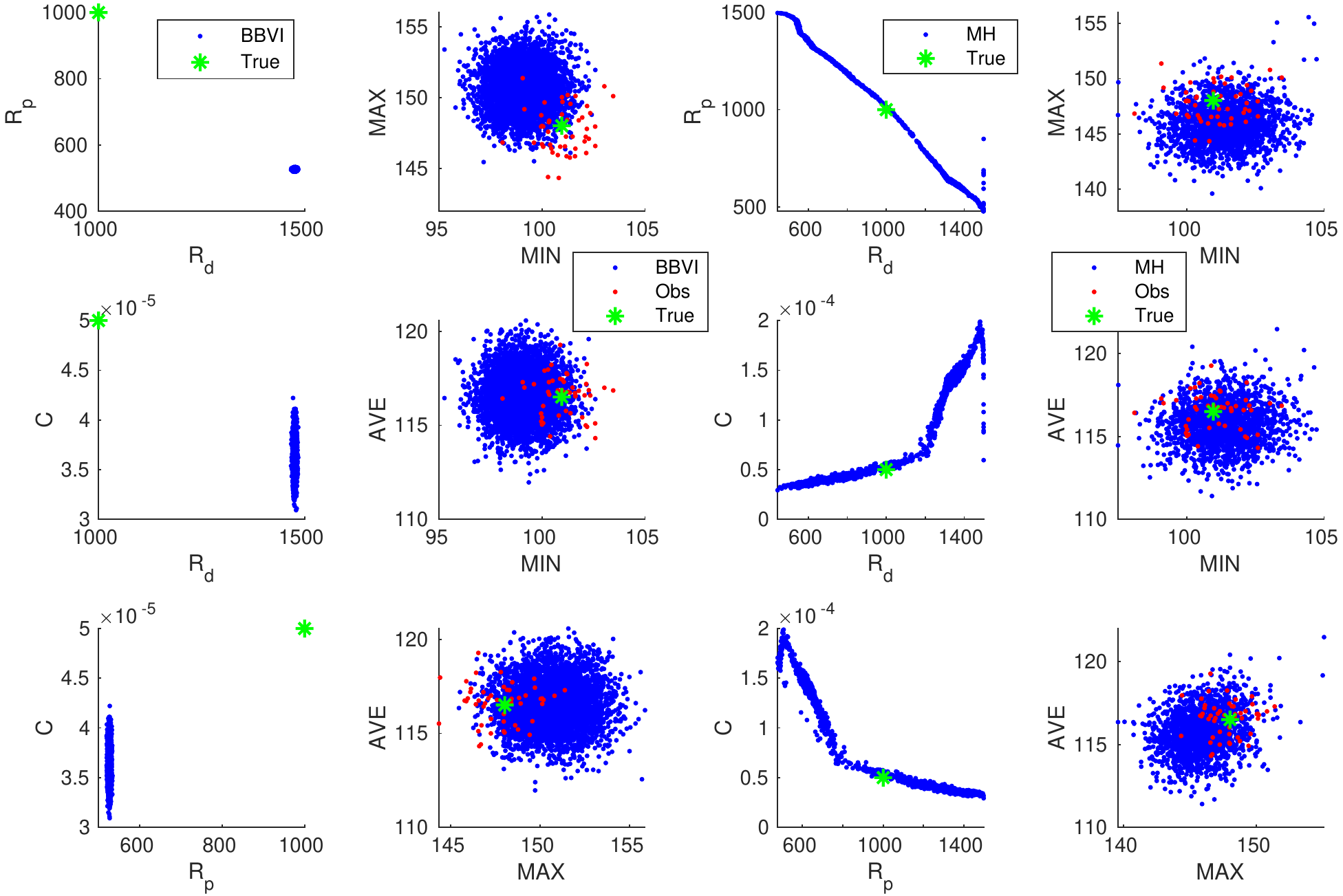}
\caption{RCR model parameters and corresponding outputs from BBVI (left two columns) and MH (right two columns) in Experiment 3. Symbols and colors are consistent with those in Figure~\ref{fig:exp2.1}.}\label{fig:RCR_BBVI_MH}
\end{figure}

The results from the BBVI and MH are shown in Figure~\ref{fig:RCR_BBVI_MH}. BBVI used a Gaussian variational distribution; $3600$ posterior samples were obtained from $4\times 10^6$ MCMC iterations with a burn-in and thinning rate of $10\%$ and $1/1000$ respectively. Since the RCR model parameters are highly correlated, it is not surprising that the mean field assumption for BBVI leads to biased posterior distributions. MH produces better results than BBVI, but still has some bias in the posterior predictive distribution, particularly in the tails. Similar to the RC model, a fixed surrogate model is used for MH, trained with a large pre-grid consisting of $20^3 = 8000$ samples.

This experiment showcases the inferential and computational superiority of NoFAS over BBVI with the mean field assumption and MH with a fixed surrogate trained from a large training dataset, for cases where there is a strong posterior correlation among parameters.
%
\subsection{Experiment 4:  Non Isomorphic Sobol Function}\label{sec:hd_model}

\noindent In this experiment, we consider a mapping from a five-dimensional parameter vector $\bm{z}\in\mathbb{R}^{5}$ onto a four-dimensional output
\begin{equation}
f(\bm{z}) = \bm{A}\,\bm{g}(e^{\bm{z}}).
\end{equation}
where $g_i(\bm{r}) = (2\cdot |2\,a_{i} - 1| + r_i) / (1 + r_i)$ with $r_i > 0$ for $i=1,\dots,5$ is the \emph{Sobol}  function~\cite{sobol2003theorems} and $\bm{A}$ is a $4\times5$ matrix. We also set
\begin{equation*}
\bm{a} = (0.084, 0.229, 0.913, 0.152, 0.826)^T \mbox{ and }\bm{A} = \frac{1}{\sqrt{2}}
\begin{pmatrix}
1 & 1 & 0 & 0 & 0\\
0 & 1 & 1 & 0 & 0\\
0 & 0 & 1 & 1 & 0\\
0 & 0 & 0 & 1 & 1\\
\end{pmatrix}.
\end{equation*}
The true parameter vector is set at $\bm{z}^{*} = (2.75,$ $-1.5, 0.25,$ $-2.5,$ $1.75)^T$. The Sobol function is bijective and analytic but $f:\mathbb{R}^{5}\to \mathbb{R}^{4}$ leads to an over-parameterized model and non-identifiability.
This is also confirmed by the fact that the curve segment $\gamma(t) = g^{-1}(g(\bm z^*) + \bm v\,t)\in Z$ gives the same model solution as $\bm{x}^{*} = f(\bm{z}^{*}) = f(\gamma(t)) \approx (1.4910,$ $1.6650,$ $1.8715,$ $1.7011)^T$ for $t \in (-0.0153, 0.0686]$, where $\bm v = (1,-1,1,-1,1)^T$. 
This is consistent with the one-dimensional null-space of the matrix $\bm A$.
Since the output $g_i(\bm r)$ of the Sobol function is $(1, 2|2a_i - 1|]$, hence $t \in \bigcap_{i=1}^5 ((1 - g_i(\bm z^*)) / v_i, (2|2a_i - 1| - g_i(\bm z^*)) / v_i] = (-0.0153, 0.0686]$.

Similar to the other 3 experiments, we generated the model output observations from a Gaussian distribution as
\begin{equation}
\bm{x} = \bm{x}^{*} + 0.01\cdot |\bm{x}^{*}| \odot \bm{x}_{0},\,\,\text{where}\,\,\bm{x}_{0} \sim \N(0,\bm I_5).
\end{equation}
The aim is to obtain posterior samples on the latent variables $\bm z$ and quantify the uncertainty given the observed data. 
The likelihood function is Gaussian and we impose a uniform prior on $\bm z$. 
The fixed surrogate model was estimated on a grid of $4^5 = 1024$ points; for NoFAS, the adaptive surrogate model was initially trained on a $3^5 = 243$ pre-grid, and then successively updated using $S_{G}=12$ samples from the $b=250$ batches computed every $c=200$ NF parameter updates. 
We used a ReLU-activated RealNVP NF with $15$ layers, where each linear masked coupling layer contains one hidden layer of $100$ nodes, and 15 batch normalization layers are added before each RealNVP layer. A RMSProp optimizer was used, with the learning rate and its exponential decay factor at $0.0005$ and $0.9999$, respectively.

\begin{figure}[!htb]
\centering
\includegraphics[width=1.0\textwidth, trim={1.5cm 1.5cm 1.5cm 1.5cm},clip]{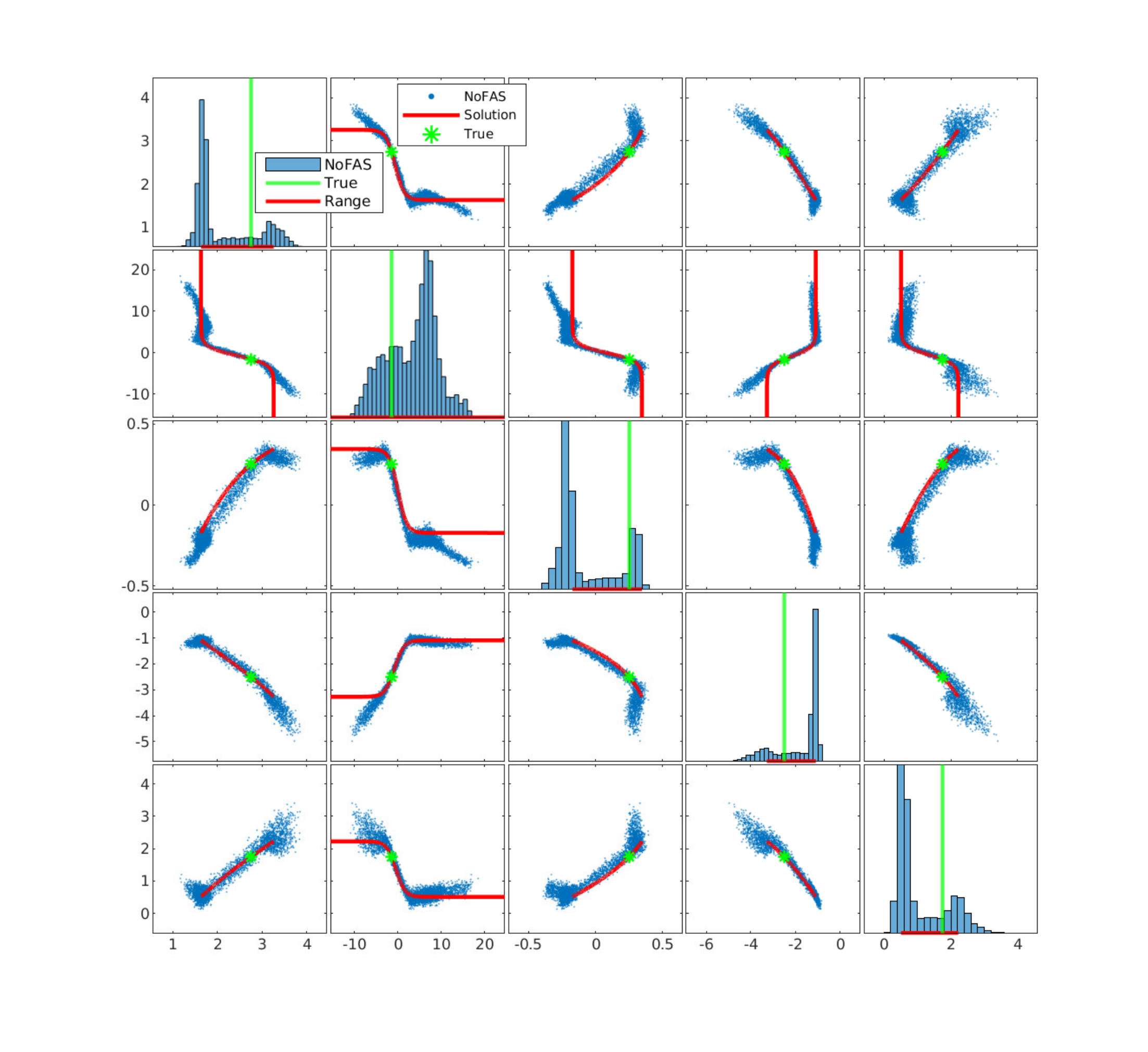}
\caption{Marginal histograms and pairwise scatter plots of the posterior samples from NoFAS in Experiment 4. The vertical green lines and the green stars represent the true parameters $\bm z^*$; the horizontal red lines indicate admissible parameter ranges compatible with the output $f(\bm{z}^{*})$, and the blue points represent the posterior samples.}\label{fig:hd_adaptive}
\end{figure}

\begin{figure}[!htb]
\centering
\includegraphics[width=0.9\textwidth, trim={1.5cm 1.5cm 1.5cm 1.5cm},clip]{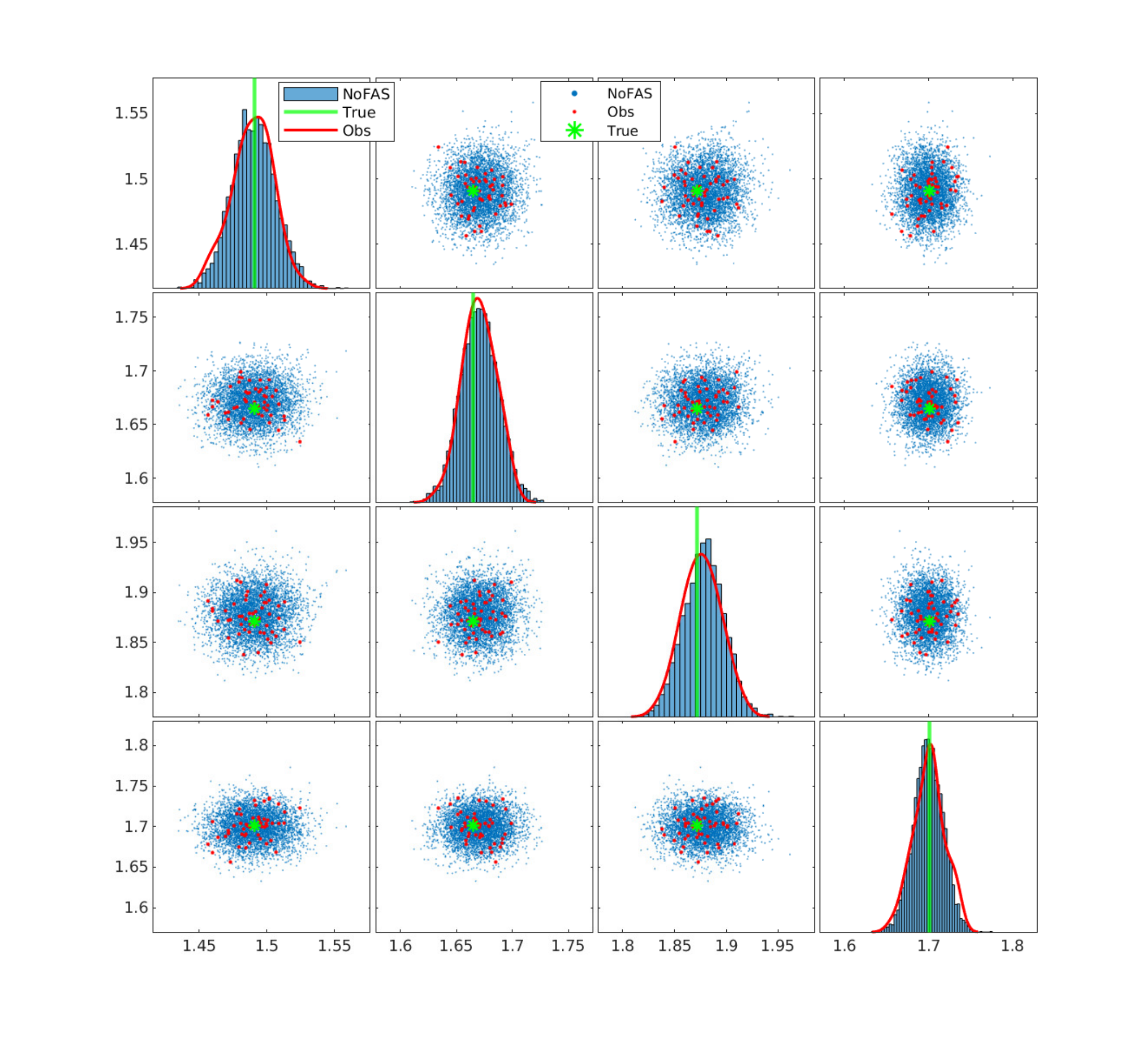}
\caption{Marginal histograms and pairwise scatter plots for the predictive posterior distribution from NoFAS in Experiment 4. The histograms on the diagonal show the marginal distributions of the posterior predictive samples for the model solutions, the green vertical lines indicate the true model solution, and the red curves are the kernel density estimation. For the non-diagonal scatter plots, the blue, red, and green dots represent the posterior predictive samples, the observations and the true model solutions, respectively.}\label{fig:4nofas}
\end{figure}

The marginal posterior histograms for the 5 model parameters and their pairwise 2-dimensional scatter plots are presented in Figure~\ref{fig:hd_adaptive}. Additionally, since the model is non-identifiable, the 5-dimensional joint posterior distribution has a \emph{ridge} $\gamma(t)$ for $t \in (-0.0153,$ $0.0686]$, highlighted in red in the pairwise plots. In the plots of $z_{2}$ vs. the other latent variables, the ridge is also characterized by a vertical or horizontal red segment, suggesting $z_2$ becomes unimportant sufficiently away from its true value $z_{2}\approx 0.223$ and making the
inference of this parameter particularly challenging. The samples from the posterior predictive distributions are presented in Figure~\ref{fig:4nofas}. The model outputs generated from the posterior predictive distributions overlap well with the actual observations. 

\begin{figure}[!htb]
\centering
\includegraphics[width=1.0\textwidth, trim={1.5cm 1.5cm 1.5cm 1.5cm},clip]{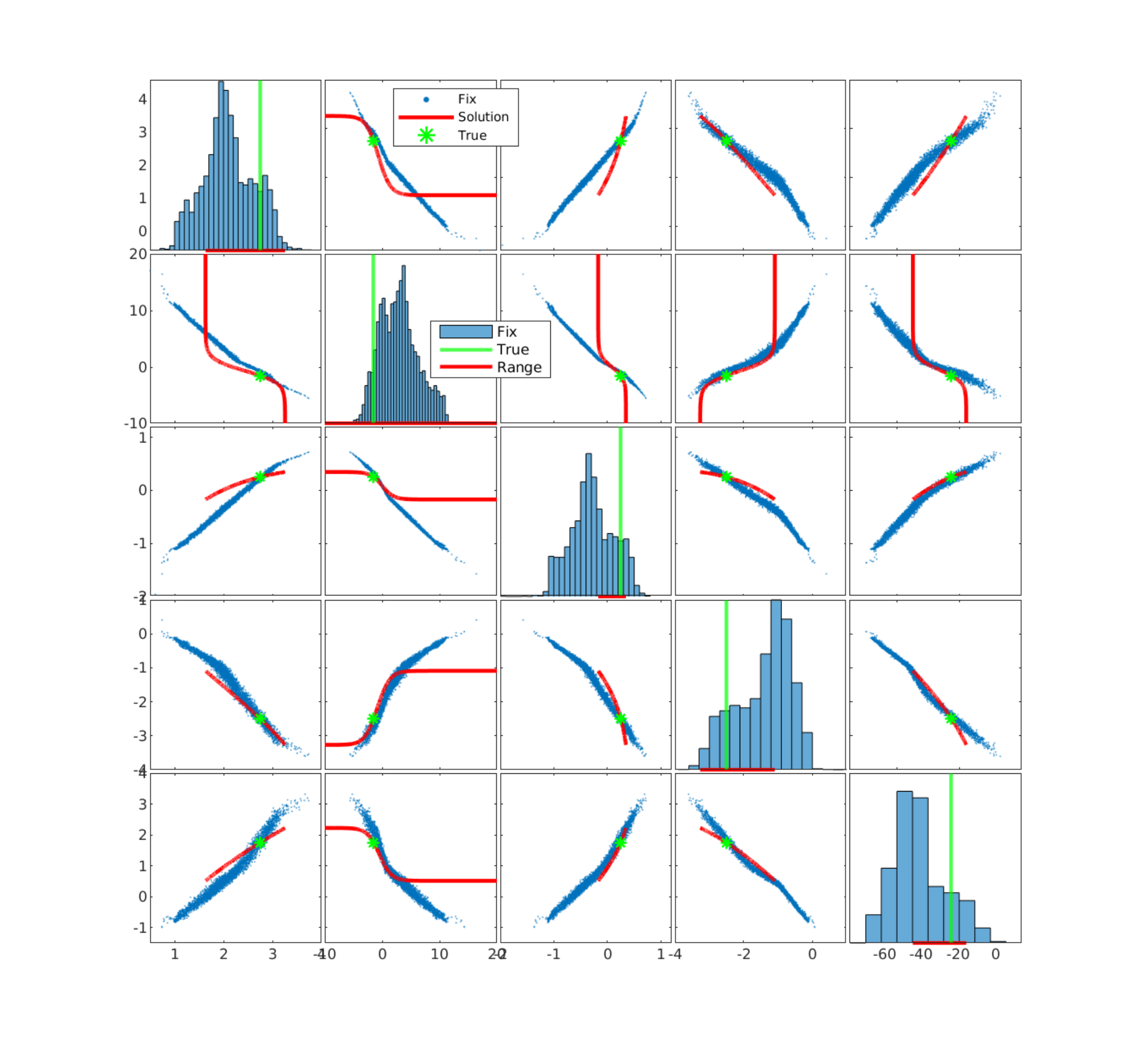}
\caption{Marginal histograms and pairwise scatter plots of the posterior samples for Experiment 4 obtained via NF with a fixed surrogate. Symbols and colors are consistent with those in Figure~\ref{fig:hd_adaptive}.}\label{fig:hd_fixed}
\end{figure}

\begin{figure}[!htb]
\centering
\includegraphics[width=0.9\textwidth, trim={1.5cm 1.5cm 1.5cm 1.5cm},clip]{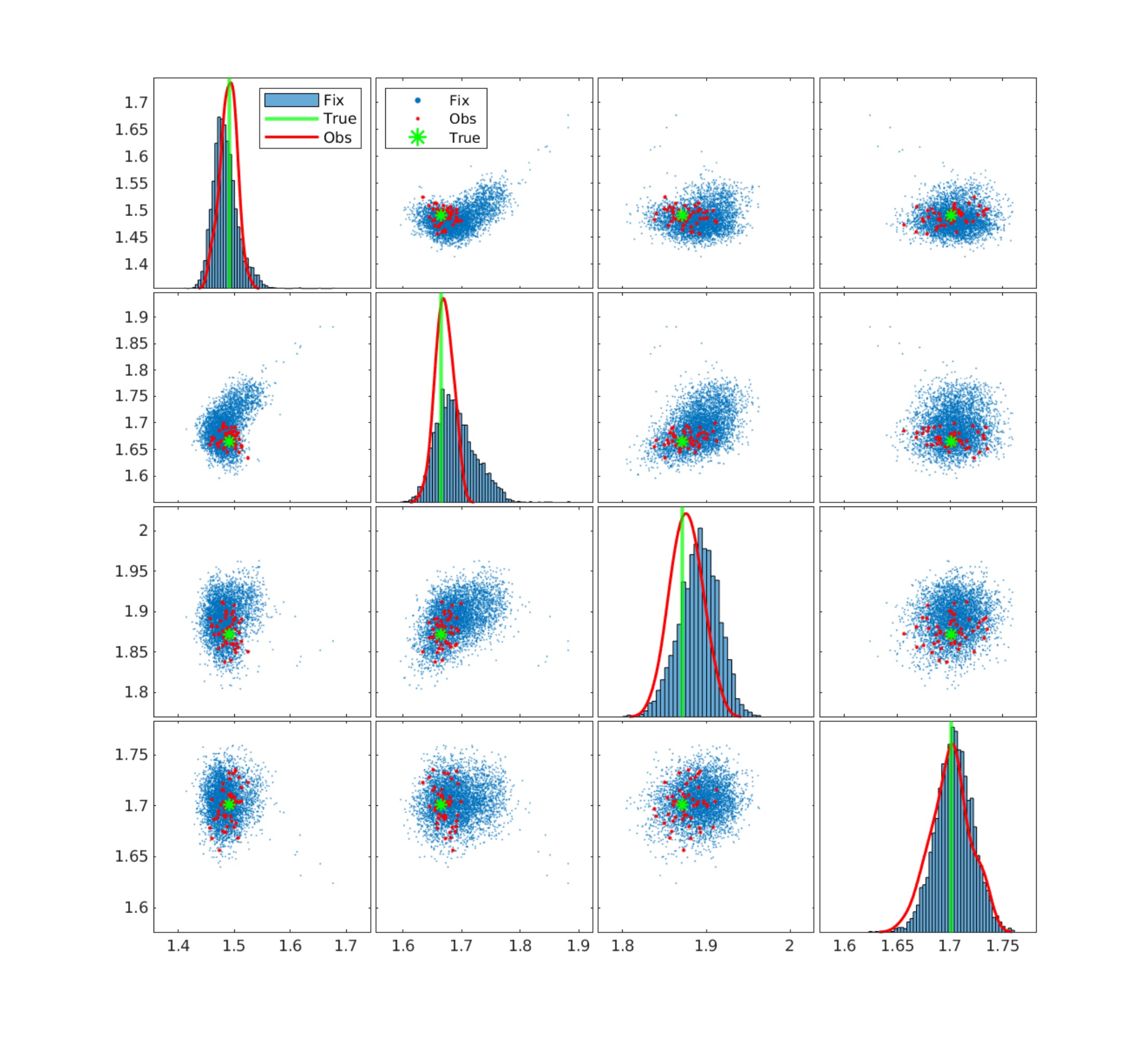}
\caption{Marginal histograms and pairwise scatter plots for the posterior predictive samples of model outputs in Experiment 4, obtained from NF with a fixed surrogate model. Symbols and colors are consistent with those in Figure~\ref{fig:4nofas}.}\label{fig:4fixed}
\end{figure}

\begin{figure}[!htb]
\centering
\includegraphics[width=1.0\textwidth, trim={1.5cm 1.5cm 1.5cm 1.5cm},clip]{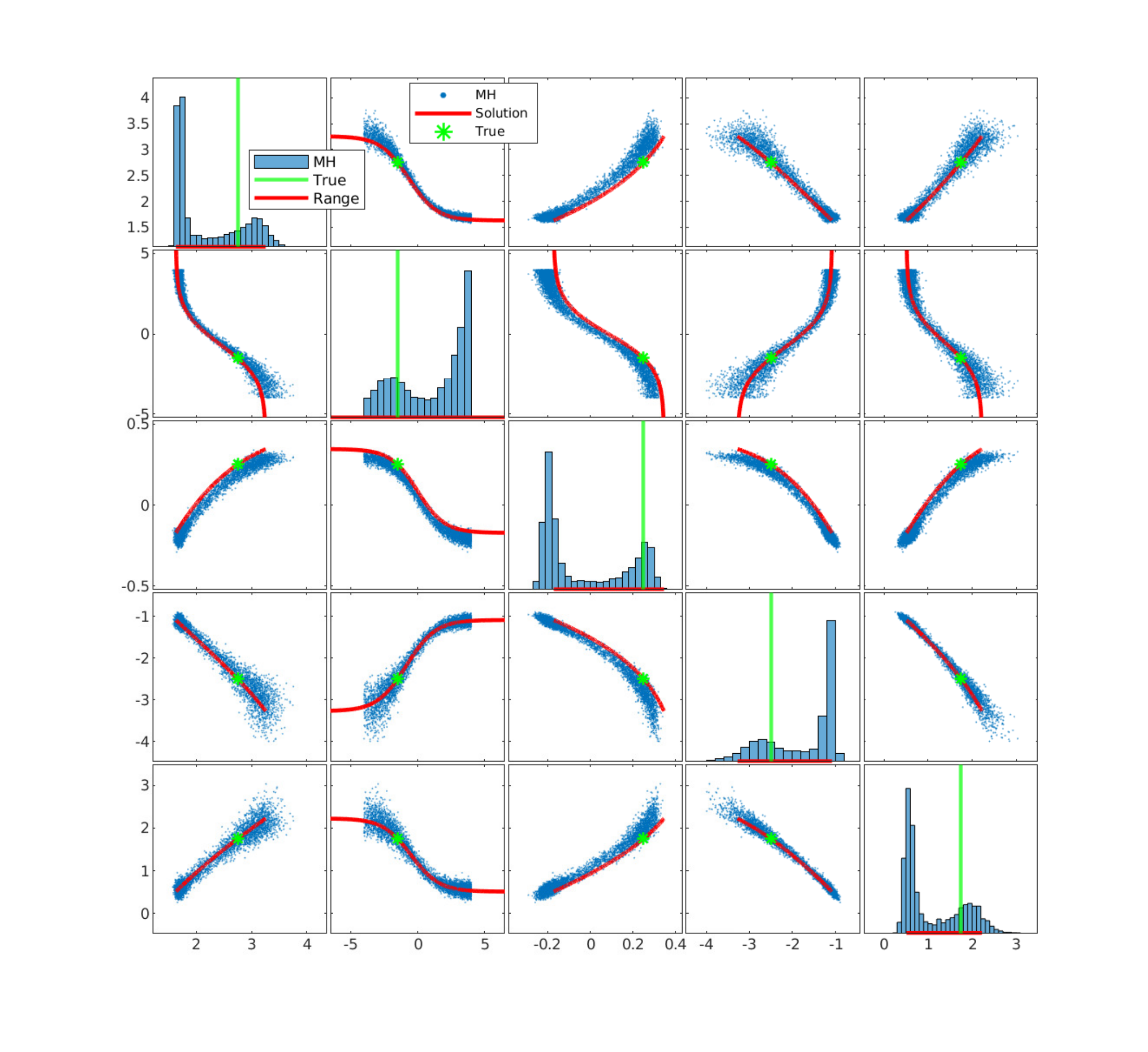}
\caption{Marginal histograms and pairwise scatter plots of the posterior samples  obtained through MH in Experiment 4. Symbols and colors are consistent with those in Figure~\ref{fig:hd_adaptive}.}\label{fig:hd_MH}
\end{figure}

\begin{figure}[!htb]
\centering
\includegraphics[width=0.9\textwidth, trim={1.5cm 1.5cm 1.5cm 1.5cm},clip]{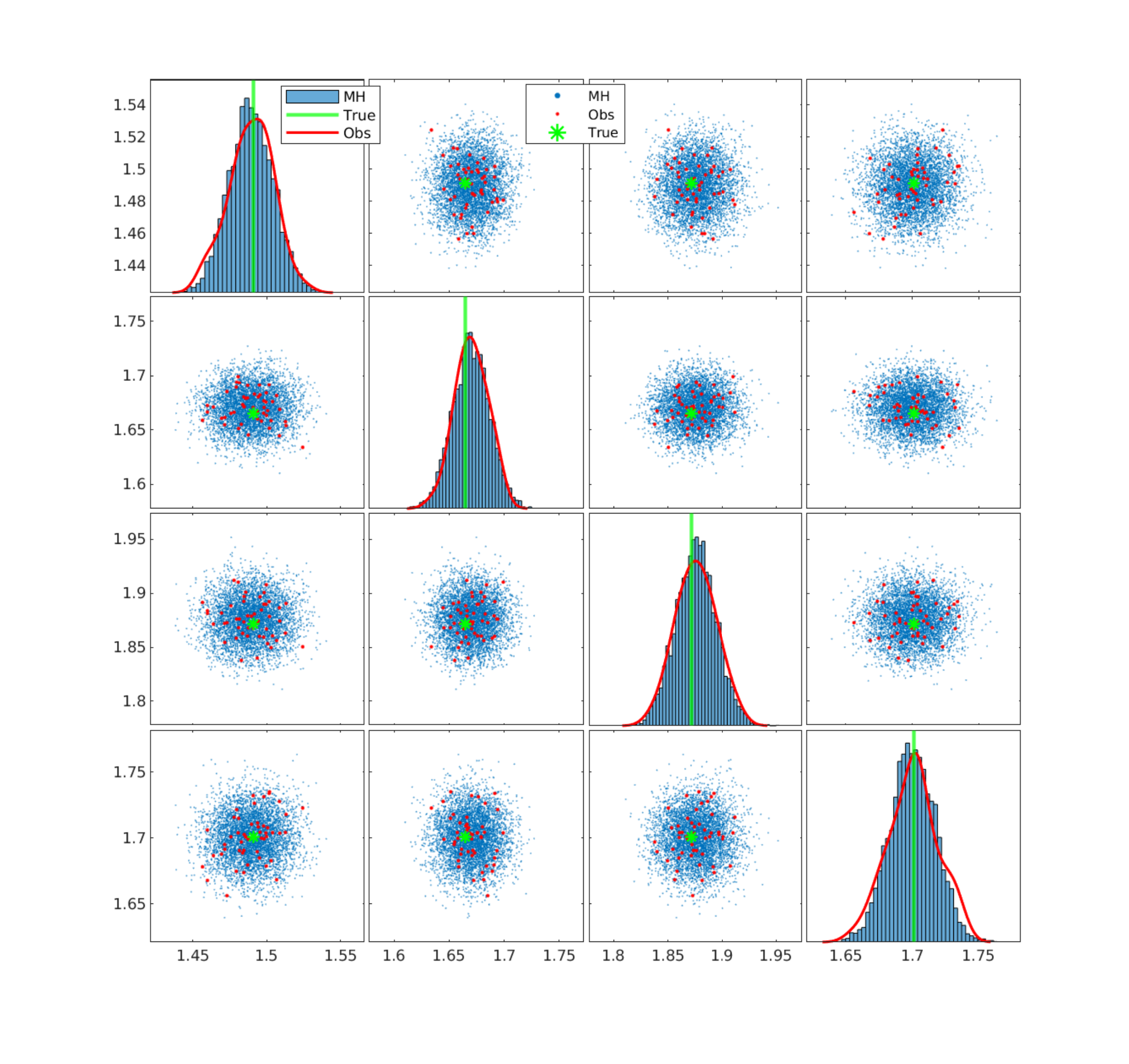}
\caption{Marginal histograms and pairwise scatter plots for the posterior predictive samples of the model solutions in Experiment 4 computed by MH. Symbols and colors are consistent with those in Figure~\ref{fig:4nofas}.}\label{fig:hd_MH_out}
\end{figure}

The results from the NF with a fixed surrogate are presented in Figure~\ref{fig:hd_fixed}. Consistent with the findings in the other experiments, the posterior samples are biased and deviate from the true parameter values. The samples from the posterior predictive distributions are presented in Figure~\ref{fig:4fixed}, which also shows some deviation of the model outputs generated from the posterior predictive distributions from the actual observations.  
We also generated posterior samples using a MH sampler. Due to the non-identifiability of the model, MH had trouble converging on parameter $z_2$ and the Markov chains for $z_{2}$ moved freely in the region where this parameter is unimportant, despite a satisfactory convergence for the other 4 parameters. 
To mitigate this problem, we constrained the prior by forcing $\bm z\in[-4,4]^5$, which leads to convergence on all parameters, as measured by the Gelman-Rubin metric~\cite{gelman1992inference}. However, the Markov chains still suffered from poor mixing and we used a burn-in of $10\%$ and a thinning interval of $1\times10^6$ to reduce the sample auto-correlation, generating $5400$ posterior samples. 
The results are presented in Figure~\ref{fig:hd_MH} and Figure~\ref{fig:hd_MH_out}. The results suggest that MH can provide parameter estimates compatible with those produced by NoFAS but might have to leverage stronger prior knowledge and requires substantially more samples for models with non-identifiable parameters.


\section{Discussion}\label{sec:disucssion}

\noindent We propose NoFAS, an approach to efficiently solve inverse problems by combining optimization-based inference with the adaptive construction of surrogate models using samples from NF. The number of forward model evaluations can be greatly reduced by adaptively training and improving the surrogate with samples from high-density posterior regions that are progressively discovered by NF. 
We also propose a flexible sampling weighting mechanism, where a number of parameter realizations from a pre-selected grid and the most recent training samples are assigned larger weights in the loss function, which significantly improves the inference results  compared to NF with uniform weights.
The trade-off between global and local surrogate accuracy can be tuned by deciding the relative budget to assign either to the pre-grid or to sample locations adaptively selected through the NF iterations.
Based on our empirical studies, the assignment of the $20\%$ to $30\%$ of the solution budget to the pre-grid produces generally accurate posterior and posterior predictive distributions, even if characterized by complex features, such as the presence of multiple modes or ridges. Assigning an excessive budget to the pre-grid can lead to poor approximation for the posterior distribution, whereas an insufficient number of pre-grid samples can cause slow convergence or even divergence.

NoFAS requires several hyperparameters to be specified or tuned, including the batch size $b$, the calibration interval $c$, and the calibration size $S_G$. 
Specifically, $b$ needs to exceed a certain threshold for NoFAS to converge. We tested different batch sizes on the RCR model and observed that the minimal loss stops decreasing for $b>50$, but $b$ needs to be at least 400 to achieve accurate uncertainty quantification.
We also tested various $c$ and found that small $c$ tends to produce too much surrogate adaptation in the early stage of NoFAS, consuming the budget too quickly or even ran out before convergence. Conversely, large $c$ could lead to inaccurate gradients, delaying the exploration of high posterior density regions. 
Note that approximation of multi-modal posterior distributions typically requires a higher $S_G$ than for uni-modal distributions. In our preliminary experiments on bimodal posterior distributions (not shown), using $S_{G}=2$ was sufficient to locate both modes.
A large $S_G$ leads to a fast consumption of the budget in the early stage of NoFAS, while a small $S_G$ could lead to an inaccurate surrogate.  We recommend $c$ and $S_G$ be jointly selected by taking the parameter space dimensionality $d$ into account. Generally speaking, problems characterized by a higher dimensionality $d$ would require a larger $b/S_G$ ratio.
As for the pre-grid weight factor $\beta_0$ and memory decay factor $\beta_1$, $\beta_0$ represents the proportion of loss computed from the pre-grid in Eqn.~\eqref{equ:surrogate_loss}, while $\beta_1$ controls the decay rate for the proportion of the loss function computed from adaptively collected samples from most to least recent. The larger $\beta_0$, the more impact the pre-grid samples will have on the results from NoFAS; the larger $\beta_1$, the shorter the memory on adaptively collected samples. We recommend values of $\beta_0 \in [0.5, 0.7]$ and $\beta_1 \in [0.01, 1.0]$ based on our experiment results.

NoFAS is designed to be agnostic with respect to the NF formulation, provided the selected flow is sufficiently expressive. In our experiments, we used RealNVP and MAF. The latter requires fewer parameters than the former to converge and has a smaller computational cost per iteration. In addition, the three stages observed for the posterior samples discussed in Section~\ref{sec:ourAlgorithm} tend to be more evident when using MAF rather than RealNVP.

\section*{Acknowledgments}

\noindent This work was supported by a NSF Big Data Science \& Engineering grant \#1918692 (PI DES), a NSF CAREER grant \#1942662 (PI DES) and used computational resources provided through the Center for Research Computing at the University of Notre Dame.

\bibliographystyle{unsrtnat}
\bibliography{Ref.bib}


\section*{Appendix}

\subsection*{A. Experiments with Black-Box Variational Inference}

\noindent Black-Box Variational Inference (BBVI)~\cite{ranganath2014black} approximates a target density $p(\bm x, \bm z)$ with a distribution $q(\bm z|\lambda)$ from a parametric family, by minimizing the Evidence Lower BOund (ELBO) defined as $\mathcal{L} = \E_{q(z|\lambda)}[\log p(\bm x, \bm z) - \log q(\bm z|\lambda)]$ (or the KL-Divergence $D(q(\bm z|\lambda)\| p(\bm x, \bm z))$) and by reducing the variance of the stochastic gradient of the ELBO by Rao-Blackwellization or control variate estimators~\cite{ranganath2014black}.

We applied BBVI on two ODE hemodynamic models (experiments 2 and 3) assuming a Gaussian variational distribution and minimizing the ELBO with RMSprop.
Results for RC model are shown in Figure~\ref{fig:RC_BBVI_MH} and RCR model in Figure~\ref{fig:RCR_BBVI_MH}. 
The BBVI estimates for the RC model appear accurate. However, this is not the case for the RCR model. Not only the correlations between $R_d$ and $R_p$ could not be captured due to the mean-field assumption in BBVI, but the posterior parameter estimates were found to be significantly biased.

\subsection*{B. Experiments with Metropolis Hastings}

\noindent Metropolis Hastings (MH) is a Markov Chain Monte Carlo (MCMC) method~\cite{gilks1995markov}, where a candidate sample $\bm{z'}$ is first generated given $\bm{z}_t$ from a pre-specified and fixed proposal distribution $q(\bm{z'}|\bm{z}_{t})$, and then accepted as $\bm{z}_{t+1} = \bm{z'}$ with probability $A(\bm{z'}, \bm{z}_t)$ $= \min(1, p(\bm{z'})$ $q(\bm{z}_t|\bm{z'})/$ $ (p(\bm{z}_t) q(\bm{z'}|\bm{z}_t)))$ or rejected ($\bm{z}_{t+1} = \bm{z}_{t}$) with probability $1- A(\bm{z'}, \bm{z}_t)$.

We applied MH in all four experiments using a multivariate Gaussian proposal distribution with a diagonal precision matrix. Additional details on the hyper-parameters and convergence metric are presented in Table~\ref{tab:MH_info}. 
The results provided by MH are shown in Figures~\ref{fig:trivial_MH}, \ref{fig:RC_BBVI_MH}, \ref{fig:RCR_BBVI_MH} and \ref{fig:hd_MH} for Experiment 1, 2, 3 and 4, respectively.
In addition, fixed surrogate models are employed for the RC and RCR models (experiments 2 and 3) to speed up the inference task. Specifically, a uniform $30\times30 = 900$ grid is used for the RC surrogate and a uniform $20\times20\times20=4000$ grid for RCR. The true model is used in Experiments 1 and 4.

In Experiment 1, MH performs as well as NoFAS but requires the computation of millions of true model solutions. Bias in the estimated parameters and model solutions is observed for the RC and RCR model respectively, likely due to the use of surrogate models.
For Experiment 4, the parameter $z_2$ becomes unimportant at a certain distance from its true value, and further changes in $z_2$ leave the posterior distribution unaltered, resulting in bad mixing and compromising the convergence of MH.
Introduction of a more informative uniform prior on $[-4,4]^5$ mitigates this problem, resulting in a similar performance to NoFAS.
\begin{table}[!ht]
\centering
\resizebox{\textwidth}{!}{
\begin{tabular}{l l l l l l l}
\toprule
Experiment  &  Var-Cov matrix  & \# of iterations & Burn-in & Thinning & Accept rate & Gelman-Rubin metric\\
\midrule
1: Closed-Form & $0.01 \ \bm I_2$ & $4\times10^6$ & 10\% & 1000 & 45.53\% & (1.0020, 1.0014)\\ 
2: RC & diag$(0.01, 0.1)$ & $2\times 10^6$ & 10\% & 1000 & 60.87\% & (0.9996, 1.0013)\\
3: RCR & $0.025 \ \bm I_3$ & $4\times 10^6$ & 10\% & 2000 & 31.42\% & (1.0605, 1.0428, 1.0307)\\
4: Sobol & $0.03 \ \bm I_5$ & $6\times 10^8$ & 10\% & 100000 & 43.23\% & (0.9984, 0.9983, 0.9982, 0.9985, 0.9983)\\
\bottomrule
\end{tabular}}
\caption{Details for the Metropolis-Hastings algorithm in the four proposed numerical experiments.}\label{tab:MH_info}
\end{table}

\subsection*{C. NoFAS Hyperparameters}

\noindent All experiments used the RMSprop optimizer and an exponential scheduler with decay factor 0.9999. All normalizing flows use ReLU activations and the maximum number of iterations is set to 25001. MADE autoencoders or linear masked coupling layers contain 1 hidden layer with 100 nodes. In addition, we use $\beta_0 = 0.5$ and $\beta_1 = 0.1$ in all experiments. The recommended values for the batch size are reported in Table~\ref{tab:Hyper_info}. Based on our experiments with different batch sizes, larger batch sizes lead to more stable results.
\begin{table}[!ht]
\centering
\resizebox{\textwidth}{!}{
\begin{tabular}{l l l l l l l l}
\toprule
Experiment & NF type & NF layers & Batch size & Budget & Updating size & Updating interval & Learning rate\\
\midrule
Closed-form & RealNVP & 5 & 200 & 64 & 2 & 1000 & 0.002\\ 
RC & MAF & 5 & 250 & 64 & 2 & 1000 & 0.003\\
RCR & MAF & 15 & 500 & 216 & 2 & 300 & 0.003\\
Sobol & RealNVP & 15 & 250 & 1023 & 12 & 250 & 0.0005\\
\bottomrule
\end{tabular}}
\caption{Hyper parameters of NoFAS used in all four experiments}\label{tab:Hyper_info}
\end{table}

\subsection*{D. Sensitivity Analysis}

\subsubsection*{D1. Pre-grid weight factor $\beta_{0}$ and memory decay factor $\beta_{1}$}

\begin{figure}[!htb]
\centering
\includegraphics[width=0.9\textwidth, trim={0.5cm 5.0cm 2.5cm 4.5cm},clip]{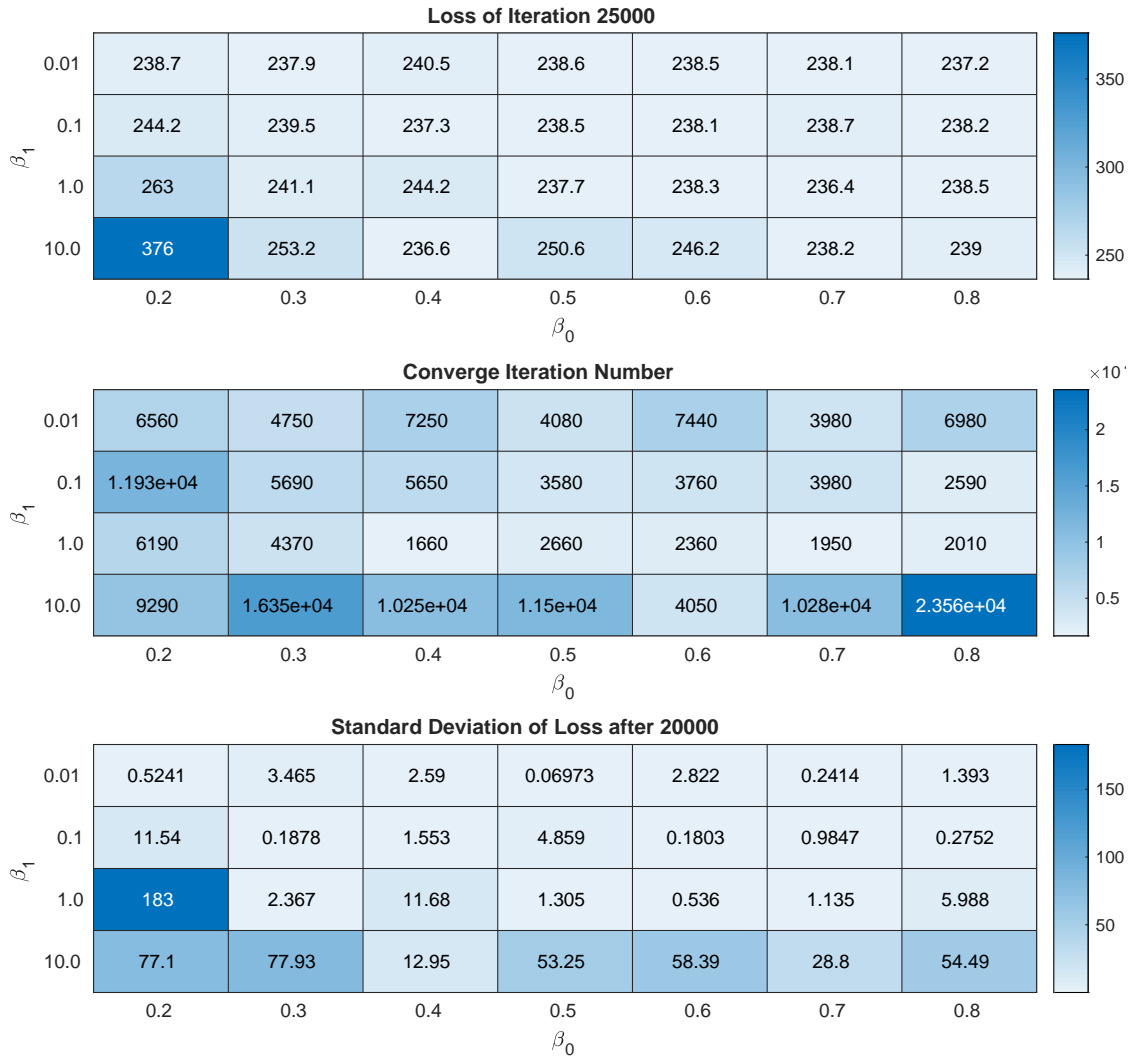}
\caption{Loss function values at iteration 25000 (top), iteration number at convergence (middle) and standard deviation of loss function values after iteration 20000 (bottom) for different combinations of $\beta_0$ and $\beta_1$ in the RCR experiment (see Section~\ref{sec:RCR}). Convergence is identified when the change in the moving average of the loss function within a 100 iteration window is less than $0.05\%$ with respect to the previous window.}\label{fig:diff_beta}
\end{figure}

\noindent We examined the performance of NoFAS varying $\beta_0$ from $0.2$ to $0.8$ and $\beta_1 \in \{0.01, 0.1, 1, 10\}$. The results are shown in Figure~\ref{fig:diff_beta}. The top heat map suggests that all the examined $\beta_0$ and $\beta_1$ values lead to similar loss function values given a sufficiently large number of iterations, but they converge with different speeds (heat map in the middle). Assigning extreme values to $\beta_0$ would slow down the convergence significantly, preventing the model to achieve sufficient local accuracy; on the other hand, too much emphasis on local regions of the parameter space would result in a surrogate model that largely ignores the global structure of $f$, possibly producing biased estimates. Using a large $\beta_1$ would put too much weight on the most recent training samples leading to slow convergence (see heat map for the convergence iteration number) and instability in the loss function (bottom heat map). In conclusion, we recommend setting $\beta_0 \in [0.5, 0.7]$ and $\beta_1 \in [0.01, 1.0]$.

\subsubsection*{D2. NF Paramater Initialization}
\begin{figure}[!htb]
\centering
\includegraphics[width=0.9\textwidth, trim={0.5cm 5.0cm 2.5cm 4.5cm},clip]{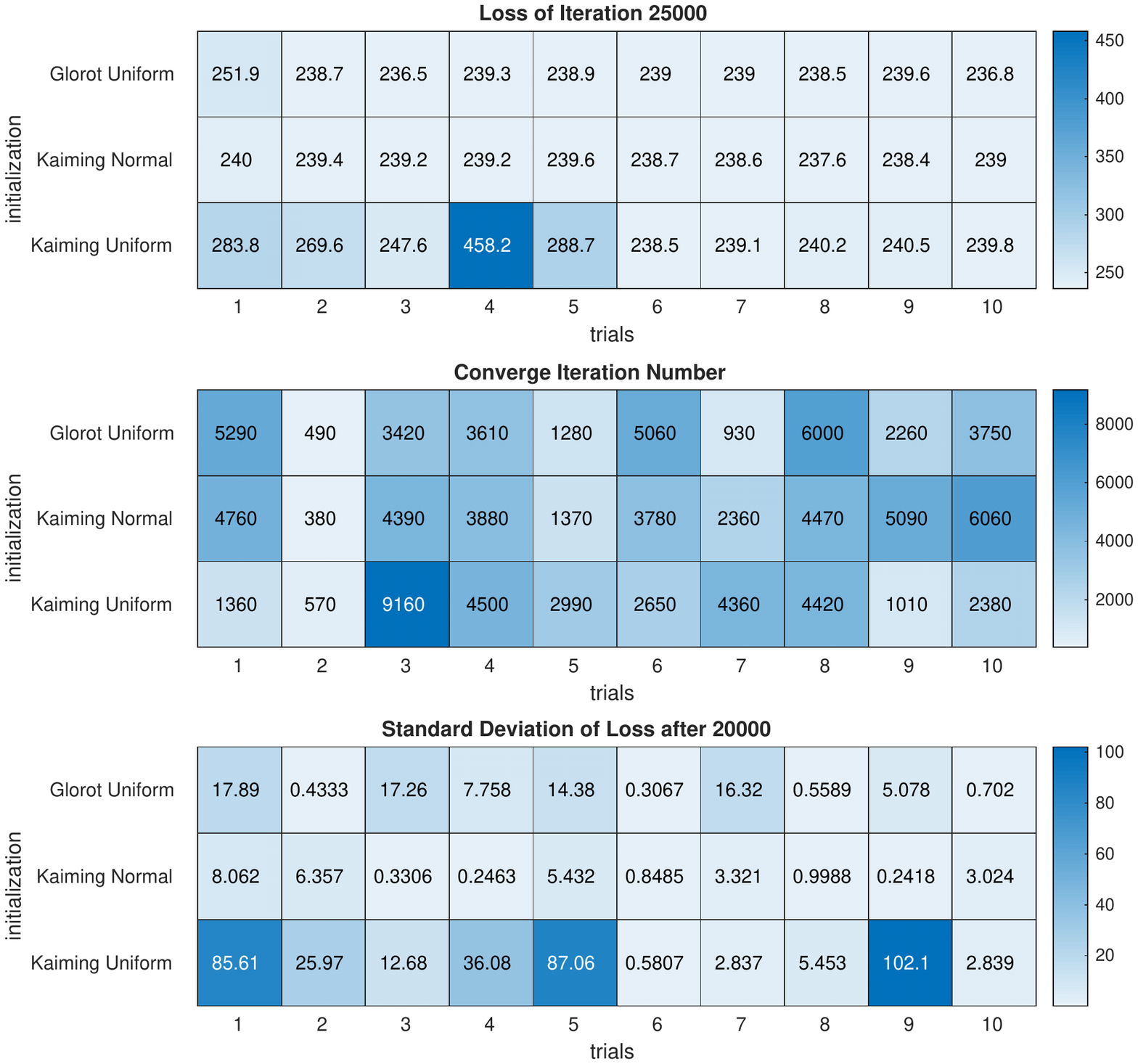}
\caption{Loss function values at iteration 25000 (top), iteration number at convergence (middle) and standard deviation of loss function values after iteration 20000 (bottom) for different weight initializations in the RCR experiment (see Section~\ref{sec:RCR}). Convergence is identified when the change in the moving average of the loss function within a 100 iteration window is less than $0.05\%$ with respect to the previous window.}\label{fig:diff_init}
\end{figure}

We used the RCR example (Section~\ref{sec:RCR}) to examine how different NF parameter initializations may affect NoFAS results. We compared the Glorot, Kaiming Uniform, and Kaiming Normal initializations in 10 repeats with different random seeds with results shown in Figure~\ref{fig:diff_init}. The Glorot initially performs slightly better than the other two initializations but overall the performance is similar for all 3 choices. Glorot and Kaiming Normal achieve slightly better quality and more stable convergence compared to Kaiming Uniform.

\end{document}